\documentclass{article}

\PassOptionsToPackage{numbers}{natbib}

\usepackage[preprint]{neurips_2021}

\usepackage[utf8]{inputenc} %
\usepackage[T1]{fontenc}    %
\usepackage{url}            %
\usepackage{booktabs}       %
\usepackage{amsfonts}       %
\usepackage{nicefrac}       %
\usepackage{microtype}      %
\usepackage{xcolor}         %

\usepackage{times}
\usepackage[pdftex]{graphicx}
\usepackage{tablefootnote}
\usepackage{longtable}
\usepackage{floatrow,wrapfig}
\usepackage{multicol}
\usepackage{enumitem}
\usepackage{algorithm,algorithmic}
\definecolor{codecomment}{HTML}{417F80}
\definecolor{causal}{HTML}{5785CD}
\definecolor{masked}{HTML}{FF9908}
\definecolor{frozen}{HTML}{3F8220}
\definecolor{citecolor}{HTML}{1B88C2}
\definecolor{refcolor}{HTML}{E01F23}
\usepackage[colorlinks=true,linkcolor=black,citecolor=citecolor,urlcolor=refcolor]{hyperref}       %
\newfloatcommand{capbtabbox}{table}[][\FBwidth]

\usepackage{amsmath,amsfonts,bm}

\def\eqref#1{equation~\ref{#1}}

\def\1{\bm{1}}

\DeclareMathAlphabet{\mathsfit}{\encodingdefault}{\sfdefault}{m}{sl}
\SetMathAlphabet{\mathsfit}{bold}{\encodingdefault}{\sfdefault}{bx}{n}

\usepackage{lipsum}
\newcommand\blfootnote[1]{%
  \begingroup
  \renewcommand\thefootnote{}\footnote{#1}%
  \addtocounter{footnote}{-1}%
  \endgroup
}

\title{Language Models as Zero-Shot Planners:\\Extracting Actionable Knowledge for Embodied Agents}

\author{Wenlong Huang\\
UC Berkeley
\And
Pieter Abbeel\\
UC Berkeley
\And
Deepak Pathak*\\
Carnegie Mellon University
\And
Igor Mordatch*\\
Google
}

\begin{document}

\maketitle

\begin{abstract}
\vspace{-0.05in}
\blfootnote{*Equal advising. Correspondence to Wenlong Huang <wenlong.huang@berkeley.edu>.}Can world knowledge learned by large language models (LLMs) be used to act in interactive environments? In this paper, we investigate the possibility of grounding high-level tasks, expressed in natural language (e.g. ``make breakfast''), to a chosen set of actionable steps (e.g. ``open fridge''). While prior work focused on learning from explicit step-by-step examples of how to act, we surprisingly find that if pre-trained LMs are large enough and prompted appropriately, they can effectively decompose high-level tasks into mid-level plans without any further training. However, the plans produced naively by LLMs often cannot map precisely to admissible actions. We propose a procedure that conditions on existing demonstrations and semantically translates the plans to admissible actions. Our evaluation in the recent VirtualHome environment shows that the resulting method substantially improves executability over the LLM baseline. The conducted human evaluation reveals a trade-off between executability and correctness but shows a promising sign towards extracting actionable knowledge from language models\blfootnote{\;Code and videos at~\url{https://huangwl18.github.io/language-planner}}.
\end{abstract}

\begin{figure}[h]
\vspace{-0.13in}
\centering
\includegraphics[width=\linewidth]{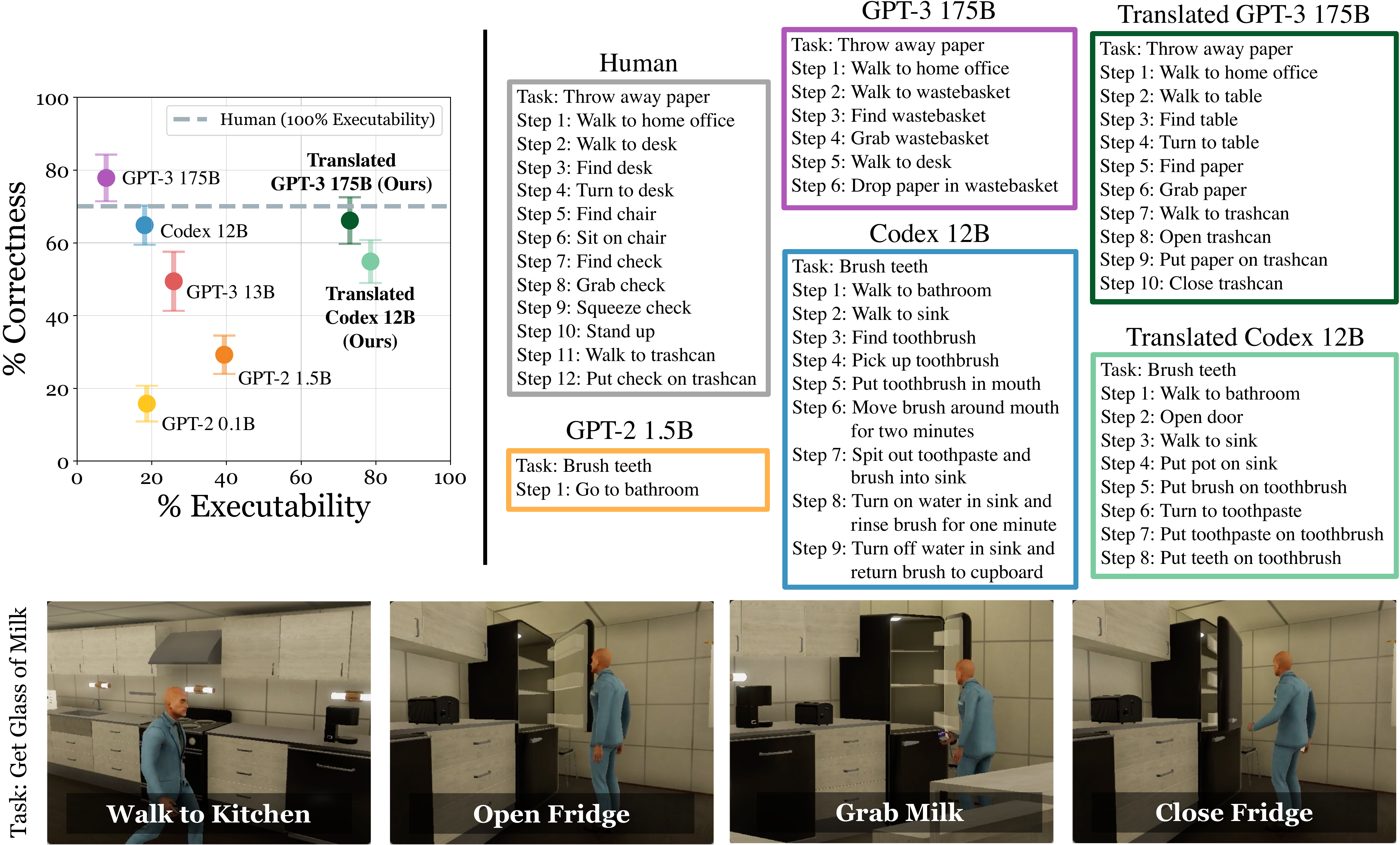}
 \caption{\small Executability v.s. semantic correctness of generated plans \textbf{(left)}, sample plans by different models \textbf{(right)}, and example environment execution (\textbf{bottom}). Large models can produce action plans indistinguishable from those by humans, but frequently are not executable in the environment. Using our techniques, we can significantly improve executability, albeit at the cost of correctness. More samples can be found in Appendix~\ref{appendix:samples}.}
 \vspace{-0.06in}
 \label{fig:results}
\end{figure}

\newpage
\hypersetup{linkcolor=black}
\tableofcontents
\hypersetup{linkcolor=refcolor}
\newpage

\section{Introduction}

Large language models (LLMs) have made impressive advances in language generation and understanding in recent years~\citep{devlin2018bert,radford2019language,raffel2019exploring,brown2020language}. See \cite{bommasani2021opportunities} for a recent summary of their capabilities and impacts.
Being trained on large corpora of human-produced language, these models are thought to contain a lot of information about the world \citep{roberts2020much,li2021implicit,bigbench} - albeit in linguistic form.

We ask whether we can use such knowledge contained in LLMs not just for linguistic tasks, but to make goal-driven decisions that can be enacted in interactive, embodied environments.
But we are not simply interested in whether we can train models on a dataset of demonstrations collected for some specific environment -- we are instead interested in whether LLMs \emph{already contain} information necessary to accomplish goals without any additional training.

More specifically, we ask whether world knowledge about how to perform high-level tasks (such as ``make breakfast'') can be expanded to a series of groundable actions (such as ``open fridge'', ``grab milk'', ``close fridge'', etc) that can be executed in the environment. For our investigation, we use the recently proposed VirtualHome environment~\citep{puig2018virtualhome}. It can simulate a large variety of realistic human activities in a household environment and supports the ability to perform them via embodied actions defined with a \texttt{verb-object} syntax. However, due to the open-ended nature of the tasks, it is difficult to autonomously evaluate their success. We rely on human evaluation (conducted on Mechanical Turk) to decide whether sequences of actions meaningfully accomplish posed tasks.

We find that large GPT-3~\citep{brown2020language} and Codex~\citep{chen2021evaluating} models, when prompted with a single fixed example of a task description and its associated sequence of actions, can produce very plausible action plans for the task we're interested in. Such completions reflect the information already stored in the model -- no model fine-tuning is involved. Additionally, we only observe this effect in the larger models. Unfortunately, despite their semantic correctness, the produced action plans are often not executable in the environment. Produced actions may not map precisely to admissible actions, or may contain various linguistic ambiguities.

We propose several tools to improve executability of the model's outputs. First, we enumerate all admissible actions and map the model's output phrases to the most semantically-similar admissible action (we use similarity measure between sentence embeddings produced by a RoBERTa model~\cite{liu2019roberta} in this work, but other choices are possible).
Second, we use the model to autoregressively generate actions in a plan by conditioning past actions that have been made admissible via the technique above. Such on-the-fly correction can keep generation anchored to admissible actions. Third, we provide weak supervision to the model by prompting the model with a known task example similar to the query task. This is somewhat reminiscent of prompt tuning approaches but does not require access to gradients or internals of the model.

Using the above tools to bias model generation, we find that we improve executability of action plans from 18\% to 79\% (see Figure \ref{fig:results}) without any invasive modifications to model parameters or any extra gradient or internal information beyond what is returned from the model's forward pass. This is advantageous because it does not require any modifications to the model training procedure and can fit within existing model serving pipelines. However, we do find there to be some drop in correctness of the action sequences generated with the above tools (as judged by humans), indicating a promising step, but requiring more research on the topic.

To summarize, our paper's contributions are as follows:
\begin{itemize}
\item We show that without any training, large language models can be prompted to generate plausible goal-driven action plans, but such plans are frequently not executable in interactive environments.
\item We propose several tools to improve executability of the model generation without invasive probing or modifications to the model.
\item We conduct a human evaluation of multiple techniques and models and report on the trade-offs between executability and semantic correctness.
\end{itemize}

\begin{figure}[t]
\centering
\includegraphics[width=\linewidth]{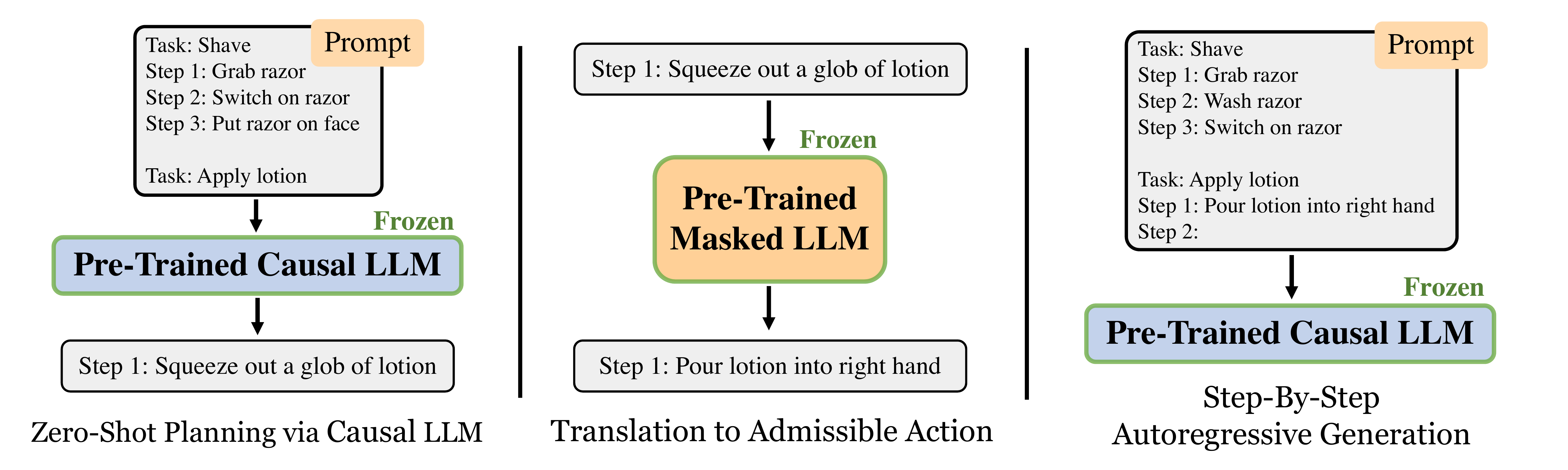}
 \caption{\small We investigate the possibility of extracting actionable knowledge from pre-trained large language models (LLMs). We first show surprising finding that \textcolor{causal}{\textbf{pre-trained causal LLMs}} can decompose high-level tasks into sensible mid-level action plans (\textbf{left}). To make the plans executable, we propose to translate each step into admissible action via another \textcolor{masked}{\textbf{pre-trained masked LLM}} (\textbf{middle}). The translated action is appended to the prompt used for generating the remaining steps (\textbf{right}). All models are kept \textcolor{frozen}{\textbf{frozen}} without additional training.}
 \label{fig:method}
\end{figure}

\section{Evaluation Framework}
Simulating open-ended tasks that resemble naturalistic human activities requires an environment to support a rich set of diverse interactions, rendering most existing embodied environments unsuitable for our investigation. One exception is VirtualHome~\citep{puig2018virtualhome}, which we evaluate on as it models complex human activities, though only in a household setting.
To measure correctness of the generated action plans, for which evaluating computationally is inherently difficult for these open-ended tasks, we conduct a human evaluation similar to~\citet{puig2018virtualhome}. We note that since no further training is involved throughout our investigations, the observations and findings presented in this paper should also translate to similar embodied environments, likely even beyond the household domain.

\subsection{Evaluated Environment: VirtualHome}
\label{sec:virtualhome}
\paragraph{Preliminaries} In VirtualHome, activities are expressed as programs. Each program consists of a sequence of textual action steps, where each step is written as:
\texttt{[action]~$\langle$arg$\rangle$(idx)}.
Each \texttt{action} refers to one of the 42 atomic actions supported in VirtualHome, such as ``walk'' and ``open''. Full list of atomic actions can be found in Appendix~\ref{appendix:mapping}. Different actions take in different numbers of \texttt{arg}, such as ``bedroom'' and ``fridge'', that are necessary for specifying an interaction. Associated with each \texttt{arg} is a unique \texttt{id} specifying the corresponding node in the environment graph, in case of multiple instances of the same object class are present in the graph. For the sake of simplicity, we omit the \texttt{id} in the remaining discussions of this paper and allow automatic assignment by the environment. An example program is shown below for the task ``Relax on sofa'':

\begin{table}[H]
\vspace{-0.05in}
  \centering
    \begin{tabular}{ll}
    &\texttt{[WALK] $\langle$living\_room$\rangle$(1)}\\
    &\texttt{[WALK] $\langle$television$\rangle$(1)}\\
    &\texttt{[FIND] $\langle$television$\rangle$(1)}\\
    &\texttt{[SWITCHON] $\langle$television$\rangle$(1)}\\
    &\texttt{[FIND] $\langle$sofa$\rangle$(1)}\\
    &\texttt{[SIT] $\langle$sofa$\rangle$(1)}\\
    &\texttt{[TURNTO] $\langle$television$\rangle$(1)}\\
    &\texttt{[WATCH] $\langle$television$\rangle$(1)}
    \end{tabular}
\end{table}

\paragraph{Evaluated Tasks} We use the \textit{ActivityPrograms} knowledge base collected by~\citet{puig2018virtualhome} for evaluation. It contains 2821 different entries annotated by Amazon Mechanical Turk (MTurk) workers. Each entry contains 1) a high-level task name (e.g. ``Watch TV''), 2) detailed instructions expressed in natural language to complete the task (e.g. ``Sit on my couch directly opposite my TV, switch on my TV with the remote control and watch''), and 3) an executable program containing all necessary steps for a robotic agent (example above). We omit the use of detailed instructions (2) as we desire direct extraction of executable programs (3) from only high-level task names (1). There are 292 distinct high-level tasks in the knowledge base, from which we randomly sample 88 held-out tasks for evaluation. The remaining 204 tasks are used as \textit{demonstration set} from which we are allowed to select as example(s) for prompting language models, or in the case of supervised fine-tuning baselines, they are used to fine-tune pre-trained language models.

\begin{algorithm}[t]
\caption{Generating Action Plans from Pre-Trained Language Models}
\label{alg:pseudocode}
\begin{algorithmic}
\STATE{\textbf{Notation Summary:}}
\STATE{$LM_P$: text completion language model (also referred as \textbf{Planning LM})}
\STATE{$LM_T$: text embedding language model (also referred as \textbf{Translation LM})}
\STATE{$\{(T_i, E_i)\}_{i=1}^{N}$: demonstration set, where $T$ is task name and $E$ is example plan for $T$}
\STATE{$C$: cosine similarity function}
\STATE{$P$: mean token log probability under $LM_P$}
\STATE \makebox[3.5em][l]{\textbf{Input:}} query task name $Q$, e.g. ``make breakfast''
\STATE \makebox[3.5em][l]{\textbf{Output:}} action plan consisting of admissible env actions, e.g. ``open fridge''
\\\hrulefill
\STATE{Extract most similar example $(T^{*}, E^{*})$ whose $T^{*}$ maximizes $C(LM_{T}(T), LM_{T}(Q))$}
\STATE{Initialize prompt with $(T^{*} + E^{*} + Q)$}
\WHILE{max step is not reached}
    \STATE{Sample $LM_P$ with current prompt to obtain $k$ single-step action phrases}
    \FOR{each sample $\hat{a}$ \textbf{and} each admissible env action $a_e$}
        \STATE{Calculate ranking score by $C(LM_{T}(\hat{a}), LM_{T}(a_e)) + \beta \cdot P(\hat{a})$}
    \ENDFOR
    \STATE{Append highest-scoring env action $a_e^{*}$ to prompt}
    \STATE{Append $a_e^{*}$ to output}
    \IF{$>50\%$ samples are 0-length \textbf{or} highest score $< \epsilon$}
        \STATE{\textbf{break}}
    \ENDIF
\ENDWHILE
\end{algorithmic}
\end{algorithm}

\subsection{Metrics}
\label{sec:metric}
A program that commands the agent to wander around in a household environment is highly executable but is mostly not correct. On the other hand, a program composed of natural language instructions annotated by humans is likely correct but cannot be executed, because its format is ambiguous and may lack necessary common-sense actions (e.g. fridge must be opened before an agent can grab things from it). We thus consider two axes for evaluation: \textbf{executability} and \textbf{correctness}.

\paragraph{Executability}
Executability measures whether an action plan can be \textit{correctly parsed} and \textit{satisfies the common-sense constraints} of the environment. To be correctly parsed, an action plan must be syntactically correct and contain only allowed actions and recognizable objects. To satisfy the common-sense constraints, each action step must not violate the set of its pre-conditions (e.g. the agent cannot grab milk from the fridge before opening it) and post-conditions (e.g. the state of the fridge changes from ``closed'' to ``open'' after the agent opens it). We report the average executability across all 88 tasks and all 7 VirtualHome scenes.

\paragraph{Correctness}
Unlike most embodied environments where the completion of a task can be easily judged, the ambiguous and multimodal nature of natural language task specification makes it impractical to obtain a gold-standard measurement of correctness\footnote{One approach could be measuring the similarity of the final environment state produced by executing predicted and human-written programs, but initial state must be kept fixed for each task, which are not appropriate for many tasks due to their open-ended nature.}. Therefore, we conduct human evaluations for the main methods. For the remaining analysis, we rely on a match-based metric that measures how similar a generated program is to human annotations. Specifically, we follow~\citet{puig2018virtualhome} and calculate the longest common subsequence (LCS) between two programs, normalized by the maximum length of the two. In the presence of multiple human-written programs for a single task, we take the maximum LCS across them. However, we note that the majority of the tasks only have one human annotation, but there are often many plausible ways to complete a certain task, making this metric imperfect at evaluation program correctness\footnote{Although LCS has a mathematical range of $[0, 1]$, we measure the LCS between different human-written programs for the same task and find an empirical maximum of $0.489$.}. Although correlation between the two is shown by~\citet{puig2018virtualhome}, we consider it only as a proxy metric in replacement of unscalable human evaluation.

\section{Method}
In this section, we investigate the possibility of extracting actionable knowledge from pre-trained language models without further training. We first give an overview of the common approach to query large language models (LLMs) and how it may be used for embodied agents in Section~\ref{sec:probe}. Then we describe an inference-time procedure that addresses several deficiencies of the LLM baseline and offers better executability in embodied environments. We break down the proposed procedure into three individual components, each discussed in Section~\ref{sec:translate},~\ref{sec:iterative},~\ref{sec:dynamic}. Pseudo-code is in Algorithm~\ref{alg:pseudocode}.

Since LMs excel at dealing with natural language text instead of the specific format required by VirtualHome as described in Section~\ref{sec:virtualhome}, we only expose natural language text to LMs. To do this, we define a bi-directional mapping for each atomic action that converts between the natural language format and the program format. For instance, ``walk to living room'' is mapped to \texttt{[\text{WALK}]~$\langle$living\_room$\rangle$(1)}. Full list of the mappings is in Appendix~\ref{appendix:mapping}.

\subsection{Querying LLMs for Action Plans}
\label{sec:probe}
Previous works have shown that large language models pre-trained on a colossal amount of data would internalize rich world knowledge that can be probed to perform various downstream tasks~\citep{radford2019language, brown2020language}. Notably, autoregressive LLMs can even perform in-context learning, an ability to solve tasks using only contextual information without gradient updates~\citep{brown2020language}. Contextual information is given as part of the input prompt and LMs are asked to complete the remaining text. It often consists of natural language instructions and/or a number of examples containing the desired input/output pairs.

We adopt the same approach to query LLMs to generate action plans for high-level tasks. Specifically, we prepend one example high-level task and its annotated action plan from the \textit{demonstration set} to the query task, as shown in Figure~\ref{fig:method}. To obtain text completion results, we sample from autoregressive LLM using temperature sampling and nucleus sampling~\citep{holtzman2019curious}. We refer to this LM as \textbf{Planning LM} and the approach using this LM for plan generation as \textbf{Vanilla \texttt{$\langle$LM$\rangle$}}, where \texttt{\textbf{$\langle$LM$\rangle$}} is replaced by specific language model such as GPT-3.

To improve the generation quality, we follow~\citet{chen2021evaluating} to sample multiple outputs for each query. However, unlike~\citet{chen2021evaluating} who investigate program synthesis and can choose the sample with highest unit test pass rate, we only consider the setting where one sample is allowed to be evaluated for each task. This is because repetitive trial-and-error
is equivalent to probing the environment for privileged information, which should not be considered viable in our setting.
For Vanilla~\texttt{$\langle$LM$\rangle$}, to choose the best action plan $X^{*}$ among $k$ samples $(X_1, X_2, ..., X_k)$, each consisting of $n_{i}$ tokens $X_i = (x_{i, 1}, x_{i, 2}, ..., x_{i, n_{i}})$, we select the sample with highest mean log probability as follows:
\begin{align}
    \operatorname*{argmax}_{X_i} \bigg( P_{\theta}(X_i) := \frac{1}{n_i} \sum_{j = 1}^{n_i} \; \text{log} \; p_{\theta} (x_{i, j} | x_{i, <j}) \bigg)
    \; \text{where} \; \theta \; \text{parameterizes the Planning LM.} \label{eq:logprob}
\end{align}

\subsection{Admissible Action Parsing by Semantic Translation}
\label{sec:translate}
One issue arises when naively following the above approach to generate action plans: the plan expressed in free-form language often cannot be mapped to unambiguous actionable steps and thus is not executable by a robotic agent.
Many reasons can cause such failures: 1) the output does not follow pre-defined mappings of any atomic action (e.g. ``I first walk to the bedroom'' is not of the format ``walk to \texttt{$\langle$PLACE$\rangle$}''), 2) the output may refer to atomic action and objects using words unrecognizable by the environment (e.g. ``microwave the chocolate milk'' where ``microwave'' and ``chocolate milk'' cannot be mapped to precise action and object), or 3) the output contains lexically ambiguous words (e.g. ``open TV'' should instead be ``switch on TV'').

Instead of developing a set of rules to transform the free-form text into admissible action steps, we propose to again leverage world knowledge learned by language models to semantically translate the action. For each admissible environment action $a_e$, we calculate its semantic distance to the predicted action phrase $\hat{a}$ by cosine similarity:
\begin{align}
    C(f(\hat{a}), f(a_e)) := \frac{f(\hat{a}) \cdot f(a_e)}{\Vert f(\hat{a}) \Vert \Vert f(a_e) \Vert}
    \; \text{where} \; f \; \text{is an embedding function.} \label{eq:cosine}
\end{align}
To embed the output action phrase and environment actions, we use a BERT-style LM~\citep{devlin2018bert, liu2019roberta} pre-trained with Sentence-BERT~\citep{reimers2019sentence} objective, to which we refer as \textbf{Translation LM}\footnote{Note that this is a different LM than the GPT-style Planning LM. Using a single LM for both purposes could as well be possible and likely more efficient, but we leave such investigation to future works.}.
The action embedding is obtained by mean-pooling the last layer hidden states across all tokens in that action phrase. While the set of admissible actions in our environment is discrete and possible to exhaustively enumerate, sampling or projection can be employed in larger discrete or continuous action spaces.

\subsection{Autoregressive Trajectory Correction} %
\label{sec:iterative}
Translating each step of the program after the entire program has been synthesized
lacks consideration of achievability of individual steps and subjects to compounding errors.
In practice, LLMs might output compounded instructions for a single step, even though it cannot be completed using one admissible action in the environment. To this end, we can instead interleave \textit{plan generation} and \textit{action translation} to allow for automatic trajectory correction. At each step, we first query Planning LM to generate $k$ samples for a single action $(\hat{a_1}, \hat{a_2}, ..., \hat{a_k})$.
For each sample $\hat{a}$, we consider both its semantic soundness and its achievability in the environment. Specifically, we aim to find admissible environment action $a_e$ by modifying the ranking scheme described in Equation~\ref{eq:logprob} as follows:
\begin{align}
    \operatorname*{argmax}_{a_e} \bigg[ \operatorname*{max}_{\hat{a}} C(f(\hat{a}), f(a_e)) + \beta \cdot P_{\theta}(\hat{a}) \bigg]
    \; \text{where} \; \beta \; \text{is a weighting coefficient.} \label{eq:ranking}
\end{align}
Then we append the translated environment action $a_e$ to the unfinished text completion. This way all subsequent steps will be conditioned on admissible actions instead of free-form action phrases generated by Planning LM. Furthermore, we can use Translation LM to detect out-of-distribution actions, those outside the capabilities of a robot, and terminate a program early instead of mapping to a faulty action. This can be achieved by setting a threshold $\epsilon$ such that if $\operatorname*{max}_{\hat{a}, a_e} C (f(\hat{a}), f(a_e)) + \beta \cdot P_{\theta}(\hat{a}) < \epsilon$ at step $t$, the program is terminated early. Since we now sample Planning LM for individual steps instead of an entire sequence, another termination condition we consider is when $>50\%$ of current-step samples are 0-length (excluding leading or trailing non-English text tokens).

\subsection{Dynamic Example Selection for Improved Knowledge Extraction}
\label{sec:dynamic}
So far in the text, we always give the same example in the prompt for all query tasks. However, consider the task of ``ordering pizza''. Prompting LLMs with this task may give the assumption that the agent is initialized in front of a computer, and the LLMs may guide the agent to search for a pizza store and click ``checkout my cart''. Although these are reasonable and feasible in the real world, such assumption cannot always be made as these interactions may not be supported in simulated environments. In fact, the closest series of actions that human experts give in VirtualHome may be ``walking to a computer'', ``switching on the computer'', and ``typing the keyboard''. Without being fine-tuned on these data, LLMs would often fail at these tasks.

To provide weak supervision at inference time, we propose to select the most similar task $T$ and its example plan $E$ from the \textit{demonstration set} to be used as the example in the prompt. Specifically, we re-use the same Translation LM introduced in Section~\ref{sec:translate} and select $(T^{*}, E^{*})$ whose high-level task name $T^{*}$ maximizes $C(f(T), f(Q))$, where $Q$ is the query task. This approach bears resemblance to several recent works~\citep{poesia2022synchromesh,gao2020making,liu2021makes,rubin2021learning}. An example is shown in Figure~\ref{fig:method} where ``Shave'' is the most similar to the query task ``Apply lotion''.

\paragraph{\textsc{Final Method}} Combining the various improvement discussed above, we refer to the final method as
\textbf{Translated \texttt{$\langle$LM$\rangle$}}
, where \textbf{\texttt{$\langle$LM$\rangle$}} is replaced by specific language model used such as GPT-3.

\begin{figure}[t]
\centering
\includegraphics[width=.9\linewidth]{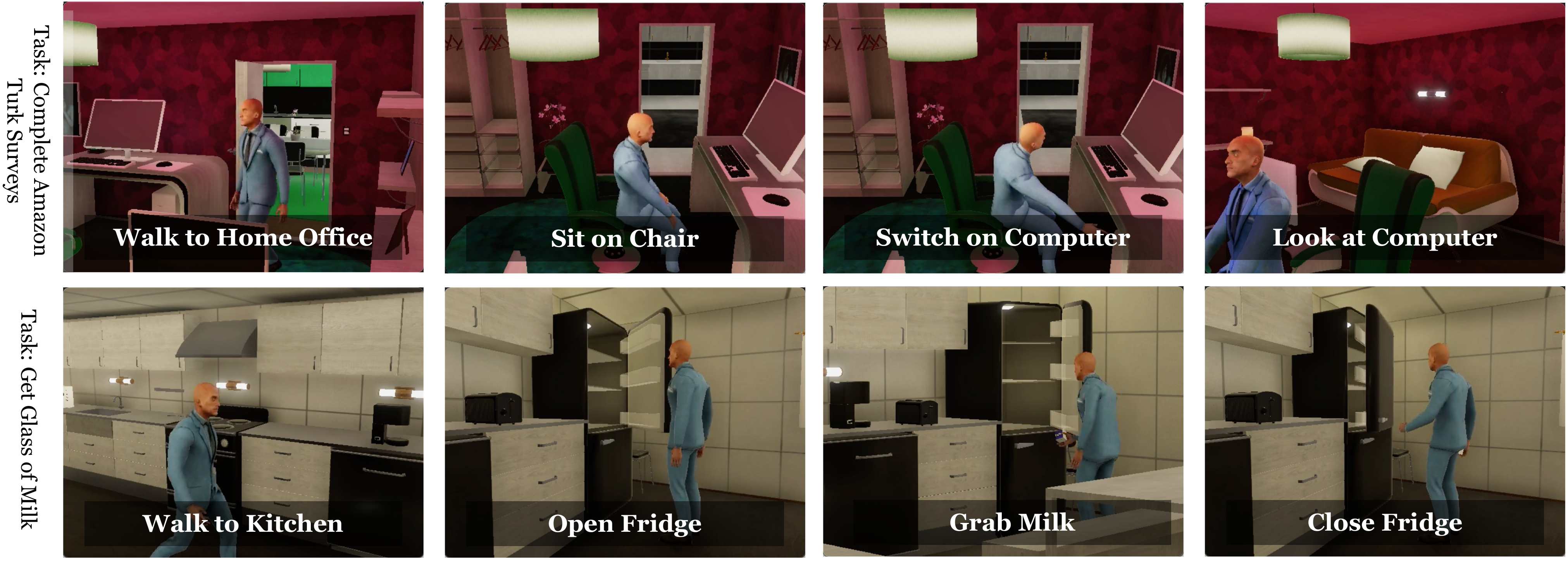}
 \caption{\small Visualization of VirtualHome programs generated by our approach. The top row shows the execution of the task ``Complete Amazon Turk Surveys'', and the bottom row shows the task ``Get Glass of Milk''. We show LLMs not only can generate sensible action plans given only high-level tasks but also contains the actionable knowledge that can be extracted for grounding in embodied environments.}
 \label{fig:virtualhome}
\end{figure}

\section{Results}
In this section, we first show that language models can generate sensible action plans for many high-level tasks, even without any additional training. Then we highlight its inadequacy when naively applied to embodied environments and demonstrate how this can be improved by again leveraging world knowledge learned by LLMs. Visualization of generated programs is shown in Figure~\ref{fig:virtualhome}.

\paragraph{Sampling from LMs} Pre-trained LMs are sensitive to sampling parameters and the specific example given in the prompt. For all evaluated methods, we perform hyperparameter search over various sampling parameters, and for methods using a fixed prompt example, we report metrics averaged across three randomly chosen examples.
To select the best run for each method, we rank the runs by the sum of LCS and executability, each normalized by human-expert scores. Further details are in Appendix~\ref{appendix:hparam}.

\paragraph{Model Choices} For Planning LM, we evaluate a representative set of causal language models. For Translation LM, we mainly use Sentence-RoBERTa-355M and provide relevant ablations in Section~\ref{sec:translation}.
GPT-3 and Codex are accessed using OpenAI API, and the remaining models are accessed through open-source packages, Hugging Face Transformers~\citep{wolf2019huggingface}
and SentenceTransformers~\citep{reimers2019sentence}, all without additional training (except for the fine-tuning baseline).

\begin{table}[t]
\begin{center}
\begin{tabular}{lllll}
\multicolumn{1}{c}{\bf Language Model} &\multicolumn{1}{c}{\bf Executability} &\multicolumn{1}{c}{\bf LCS}  &\multicolumn{1}{c}{\bf Correctness}\\
\midrule
Vanilla GPT-2 117M & 18.66\% & 3.19\% & 15.81\% (4.90\%) \\
Vanilla GPT-2 1.5B & 39.40\% & 7.78\% & 29.25\% (5.28\%) \\
Vanilla Codex 2.5B & 17.62\% & 15.57\% & 63.08\% (7.12\%) \\
Vanilla GPT-Neo 2.7B & 29.92\% & 11.52\% & 65.29\% (9.08\%) \\
Vanilla Codex 12B & 18.07\% & 16.97\% & 64.87\% (5.41\%) \\
Vanilla GPT-3 13B & 25.87\% & 13.40\% & 49.44\% (8.14\%) \\
Vanilla GPT-3 175B & 7.79\% & 17.82\% & 77.86\% (6.42\%) \\
\midrule
Human & 100.00\% & N/A & 70.05\% (5.44\%) \\ %
Fine-tuned GPT-3 13B & 66.07\% & 34.08\% & 64.92\% (5.96\%) \\
\midrule
\textsc{\textbf{\textcolor{gray}{Our Final Methods}}}\\
Translated Codex 12B & 78.57\% & 24.72\% & 54.88\% (5.90\%) \\ %
Translated GPT-3 175B & 73.05\% & 24.09\% & 66.13\% (8.38\%) %
\end{tabular}
\caption{\small Human-evaluated correctness and evaluation results in VirtualHome. Although action plans generated by large language models can match or even surpass human-written plans in correctness measure, they are rarely executable. By translating the naive action plans, we show an important step towards grounding LLMs in embodied environments, but we observe room to achieve this without trading executability for correctness.  We also observe a failure mode among smaller models that lead to high executability. For correctness measure, standard error of the mean across 10 human annotators is reported in the parenthesis.}
\label{tab:main}
\end{center}
\end{table}

\subsection{Do LLMs contain actionable knowledge for high-level tasks?}
\label{sec:vanilla-correctness}
We first investigate whether LLMs can generate sensible action plans expressed in free-form language. We use the approach described in Section~\ref{sec:probe} to query pre-trained LLMs. To evaluate the correctness of generated action plans, we conduct human evaluations. 
For each model, we ask 10 human annotators to determine -- by answering ``Yes'' or ``No'' -- whether each task can be completed using provided action steps. To provide a reference of how humans might rate the action plans provided by other humans, we also ask annotators to rate the human-written action plans included in the VirtualHome dataset for the same set of tasks. In contrast to the free-form text output by LLMs, humans wrote the plans using a graphical programming interface that enforces strict syntax and a chosen set of atomic action vocabulary, which limit the expressivity and the completeness of their answers\footnote{~\citet{puig2018virtualhome} also conduct a human evaluation on 100 randomly sampled human-written programs and show that $64\%$ of them are complete (i.e. contain all necessary steps). Readers are encouraged to refer to~\citet{puig2018virtualhome} for a more comprehensive analysis of the dataset.}. More details of our human evaluation procedure can be found in Appendix~\ref{appendix:human}.

We show the human evaluation results in Figure~\ref{fig:results}, where the y-axis shows correctness averaged across all tasks and all annotators. Surprisingly, when LLMs are large enough and without imposed syntactic constraints, they can generate highly realistic action plans whose correctness -- as deemed by human annotators -- even surpasses human-written action plans.
We also observe some level of correctness for smaller models such as GPT-2. However, inspection of its produced output indicates that it often generates shorter plans by ignoring common-sense actions or by simply rephrasing the given task (e.g. the task ``Go to sleep'' produces only a single step ``Go to bed''). These failure modes sometimes mislead human annotators to mark them correct as the annotators may ignore common-sense actions in their judgment as well, resulting in a higher correctness rate than the quality of the output shows.

\subsection{How executable are the LLM action plans?}
\label{sec:vanilla-exec}
We analyze the executability of LLM plans by evaluating them in all 7 household scenes in VirtualHome. As shown in Table~\ref{tab:main}, we find action plans generated naively by LLMs are generally not very executable. Although smaller models seem to have higher executability, we find that the majority of these executable plans are produced by ignoring the queried task and repeating the given example of a different task. This is validated by the fact that smaller models have lower LCS than larger models despite having high executability, showing that this failure mode is prevalent among smaller models. In contrast, larger models do not suffer severely from this failure mode. Yet as a result of being more expressive, their generated programs are substantially less executable. 

\subsection{Can LLM action plans be made executable by proposed procedure?}
\label{sec:translated-exec}
We evaluate the effectiveness of our proposed procedure of action translation. We first create a bank of all allowed 47522 action steps in the environment, including all possible combinations of atomic actions and allowed arguments/objects. Then we use an off-the-shelf Sentence-RoBERTa~\citep{liu2019roberta, reimers2019sentence} as Translation LM to create embeddings for actions and output text. For better computational efficiency, we pre-compute the embeddings for all allowed actions, leaving minor computation overhead for our procedure over the baseline methods at inference time. As shown in Table~\ref{tab:main}, executability of generated programs is significantly improved. Furthermore, we also observe improved LCS because the translated action steps precisely follow the program syntax and thus are more similar to the plans produced by human experts. Sample output is shown in Figure~\ref{fig:results} and a larger random subset of generated samples can be found in Appendix~\ref{appendix:samples}. 

To validate their correctness, we again perform human evaluations using the same procedure from Section~\ref{sec:vanilla-correctness}. Results are shown in Table~\ref{tab:main}. We find that despite being more similar to human-written plans as they follow strict syntax, the programs are deemed less correct by humans compared to their vanilla counterparts. By examining the output, we observe two main sources of errors. First, we find Translation LM is poor at mapping compounded instructions to a succinct admissible action, e.g. ``brush teeth with toothbrush and toothpaste''.
Second, we find that the generated programs are sometimes terminated too early. This is partly due to the imperfect expressivity of the environment; certain necessary actions or objects are not implemented to fully achieve some tasks, so Translation LM cannot map to a sufficiently similar action. This is also reflected by our human evaluation results of the programs written by other humans, as only 70\% of the programs are considered complete.

\section{Analysis and Discussions}

\subsection{Ablation of design decisions}
We perform ablation studies for the three components of our proposed procedure, described in Section~\ref{sec:translate},~\ref{sec:iterative}, and~\ref{sec:dynamic} respectively. As shown in Table~\ref{tab:ablation}, leaving out any of the three components would all lead to decreased performance in both executability and LCS. An exception is Translated GPT-3 w/o Trajectory Correction, where we observe a slight improvement in LCS at the expense of a considerable drop in executability. Among the three proposed components, leaving out action translation leads to the most significant executability drop, showing the importance of action translation in extracting executable action plans from LLMs.

\begin{table}[H]
\centering
\begin{tabular}{lll}
\multicolumn{1}{c}{\bf Methods} &\multicolumn{1}{c}{\bf Executability} &\multicolumn{1}{c}{\bf LCS} \\
\midrule
Translated Codex 12B & \bf{78.57\%} & \bf{24.72\%} \\
- \textbf{w/o} Action Translation & 31.49\% & 22.53\% \\
- \textbf{w/o} Dynamic Example & 50.86\% & 22.84\% \\
- \textbf{w/o} Trajectory Correction & 55.19\% & 24.43\%\\
\midrule
Translated GPT-3 175B & \bf{73.05\%} & 24.09\% \\
- \textbf{w/o} Action Translation & 36.04\% & 24.31\% \\
- \textbf{w/o} Dynamic Example & 60.82\% & 22.92\% \\
- \textbf{w/o} Trajectory Correction & 40.10\% & \bf{24.98}\%
\end{tabular}
\caption{Ablation of three proposed techniques.}
\label{tab:ablation}
\end{table}

\subsection{Are the generated action plans grounded in the environment?}
\label{sec:grounding}
\begin{figure}[H]
  \vspace{-0.15in}
\centering
\includegraphics[width=0.75\linewidth]{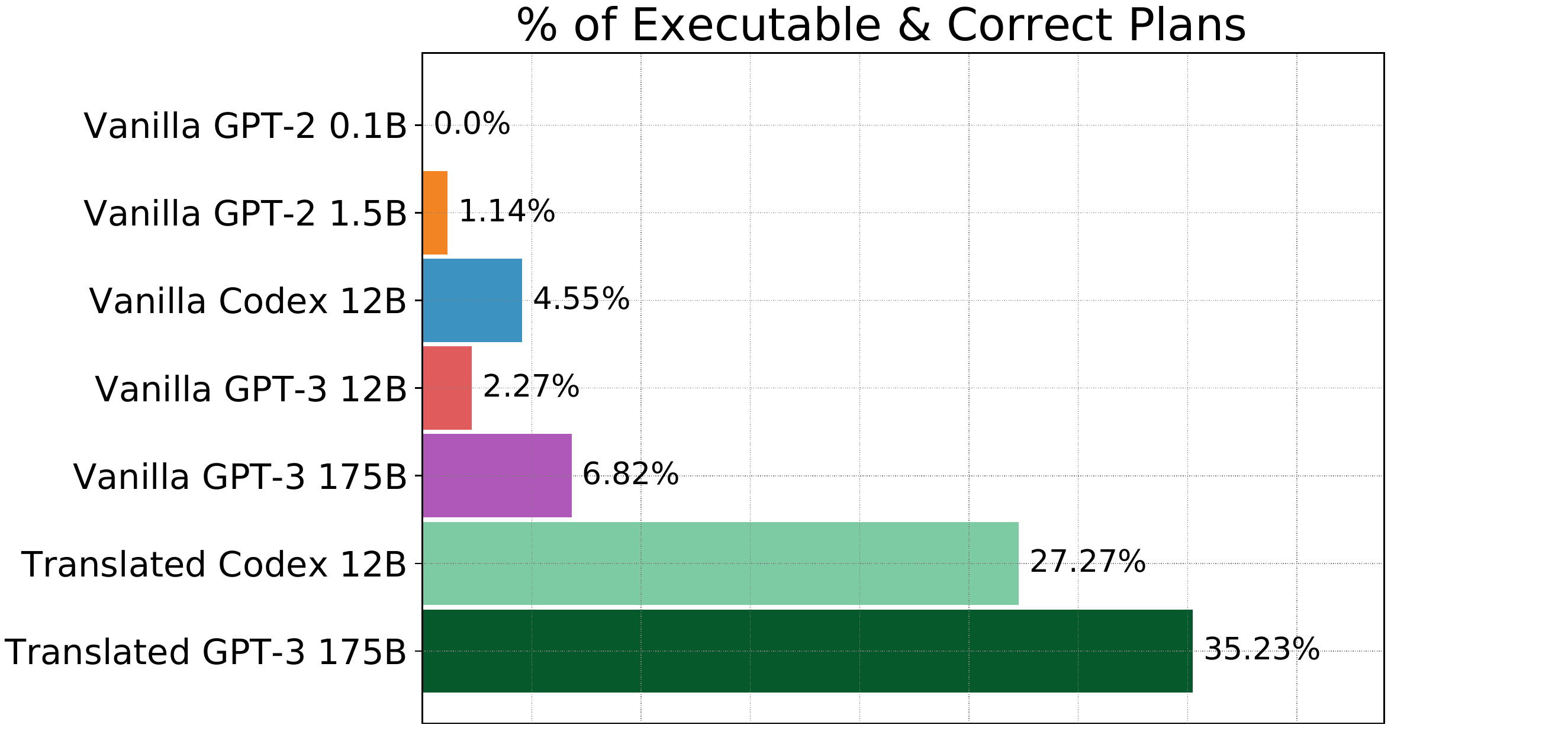}
 \caption{Percentage of both executable and correct action plans generated by LMs.}
 \label{fig:grounded}

\end{figure}

Since successful execution of correct action plans directly measures grounding, we calculate the percentage of generated action plans that are both \textit{correct} and \textit{executable}. We deem an action plan to be correct if 70\% or more human annotators decide it is correct. Human-written plans are 100\% executable, of which 65.91\% are deemed correct. Results for LMs are shown in Figure~\ref{fig:grounded}.

Although smaller LMs such as GPT-2 can generate highly executable action plans as shown in Table~\ref{tab:main}, these executable plans mostly are not correct, as they often repeat the given example or do not contain all necessary steps. Increasing model parameters can lead to some improvement in generating plans that are both executable and correct, yet it scales poorly with the parameter count. In the meantime, action translation offers a promising way towards grounding actionable knowledge by producing executable and correct plans, though a large gap remains to be closed to reach human-level performance (65.91\%).

\subsection{Effect of Different Translation LMs}
\label{sec:translation}
In this section, we study the effect of using different Translation LM. We compare two size variants of Sentence BERT and Sentence RoBERTa~\citep{devlin2018bert,liu2019roberta,reimers2019sentence} trained on the STS benchmark~\citep{cer2017semeval} and a baseline using averaged GloVe embeddings~\citep{pennington2014glove}. Results are shown in Table~\ref{tab:translation}. Notably, we do not observe significant differences in executability and LCS across different variants of BERT and RoBERTa. We hypothesize that this is because any language models trained on reasonably large datasets should be capable of the single-step action phrase translation considered in this work. However, simply using average GloVe embeddings would lead to significantly reduced performance.

\begin{table}[H]
\begin{center}
\begin{tabular}{lllll}
\multicolumn{1}{c}{\bf Translation LM} &\multicolumn{1}{c}{\bf Parameter Count} &\multicolumn{1}{c}{\bf Executability} &\multicolumn{1}{c}{\bf LCS} \\
\midrule
\textcolor{gray}{\textsc{\textbf{Codex 12B as Planning LM}}}\\
Avg. GloVe embeddings & - & 46.92\% & 9.71\% \\
Sentence Bert (base) & 110M & 73.21\% & 24.10\% \\
Sentence Bert (large) & 340M & 75.16\% & 20.79\% \\
Sentence RoBERTa (base) & 125M & 74.35\% & 22.82\% \\
Sentence RoBERTa (large) & 325M & \bf{78.57\%} & \bf{24.72\%}\\
\midrule
\textcolor{gray}{\textsc{\textbf{GPT-3 175B as Planning LM}}}\\
Avg. GloVe embeddings & - & 47.40\% & 12.16\% \\
Sentence Bert (base) & 110M & \textbf{77.60\%} & \textbf{24.49\%} \\
Sentence Bert (large) & 340M & 67.86\% & 21.24\% \\
Sentence RoBERTa (base) & 125M & 72.73\% & 23.64\% \\
Sentence RoBERTa (large) & 325M & 73.05\% & 24.09\%
\end{tabular}
\caption{Effect of different Translation LMs on executability and LCS.}
\label{tab:translation}
\end{center}
\end{table}

\subsection{Can LLMs generate actionable programs by following step-by-step instructions?}
\begin{figure}[H]
\vspace{-.03in}
\begin{floatrow}
\ffigbox{%
  \includegraphics[width=\linewidth]{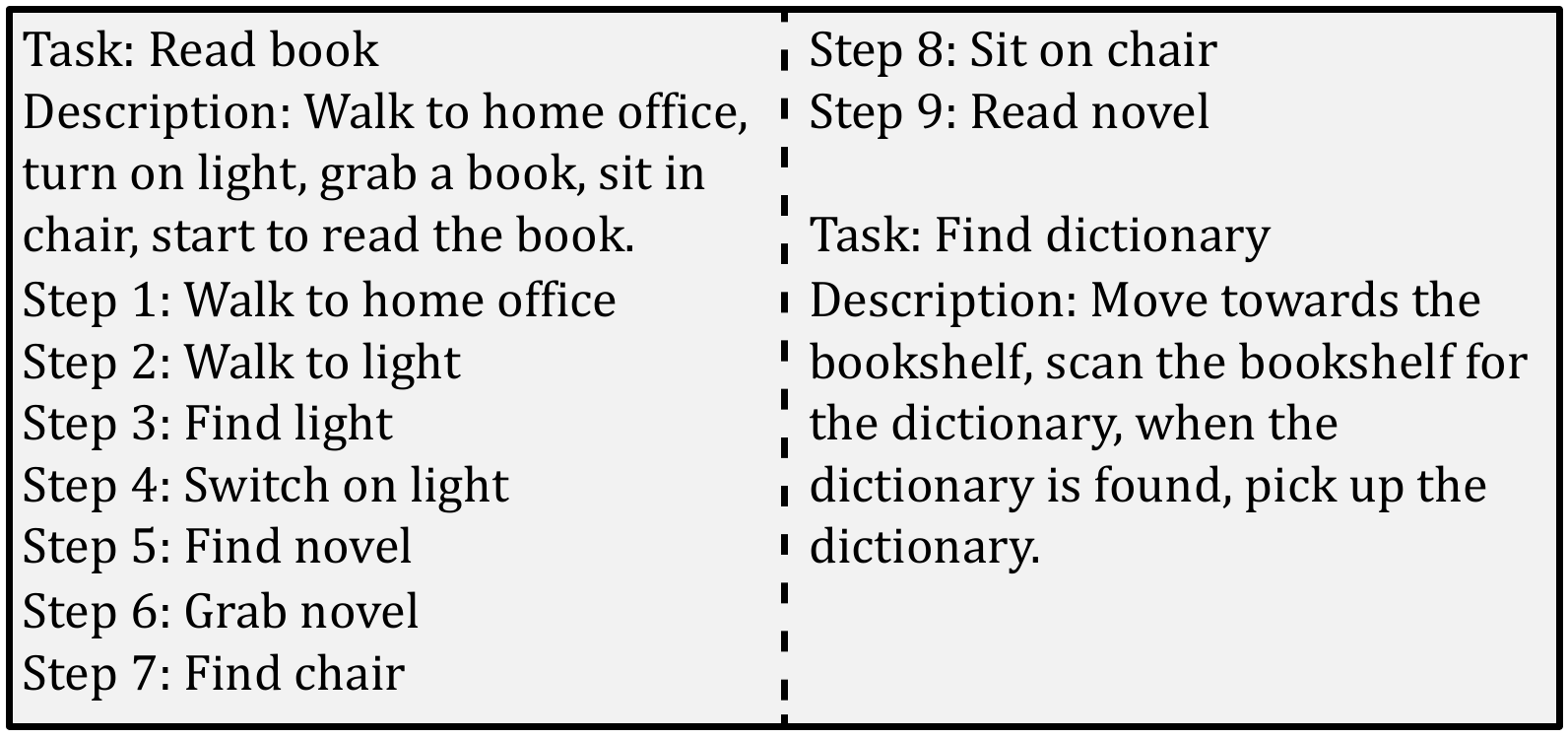}
}{%
  \caption{An example prompt containing step-by-step instructions.}\label{fig:desc}
}
\capbtabbox{%
\resizebox{\linewidth}{!}{%
  \begin{tabular}{lll}
    \multicolumn{1}{c}{\bf Methods} &\multicolumn{1}{c}{\bf Executability} &\multicolumn{1}{c}{\bf LCS} \\
    \midrule
    Translated Codex 12B & 78.57\% & 32.87\% \\
    Translated GPT-3 175B & 74.15\% & 31.05\% \\
    \midrule
    Supervised LSTM & - & 34.00\% \\
    \\
    \\
    \end{tabular}}
}{%
  \caption{Executability and LCS when conditioned on step-by-step instructions.}\label{tab:desc}
}
\end{floatrow}
\end{figure}

Prior works often focus on translating step-by-step instructions into executable programs.
Specifically, instead of only providing a high-level task name, \textit{how-to} instructions are also provided, as shown in Figure~\ref{fig:desc}.
Although this setting is easier as it does not require rich prior knowledge, \textit{how-to} instructions can help resolve much ambiguity of exactly how to perform a high-level task when multiple solutions are possible.
To investigate whether pre-trained LLMs are capable of doing this without additional training, we include these instructions in the prompt and evaluate LLMs with the proposed procedure.
We compare to a supervised baseline from VirtualHome that trains an LSTM~\citep{hochreiter1997long} from scratch on human-annotated data.
Since the code to train the baseline is not publicly released and a different train/test split is likely used, we only show results reported in~\citet{puig2018virtualhome} as a crude reference. We also cannot compare executability as it is not reported.
Results are shown in Table~\ref{tab:desc}. Surprisingly, without being fine-tuned on any domain data, Translated Codex/GPT-3 can attain LCS close to supervised methods while generating highly executable programs.

\vspace{0.1in}
\subsection{Analysis of program length}
Shorter programs have a natural advantage of being more executable as they need to satisfy less pre/post-conditions, albeit being prone to incompleteness. To validate the proposed approach does not simply generate very short programs, we calculate the average program length across the 88 evaluated tasks. Results are shown in Table~\ref{tab:length}. Mirroring the observations made in Section~\ref{sec:vanilla-correctness} and Section~\ref{sec:vanilla-exec}, we find smaller LMs such as GPT-2 tend to generate shorter programs than larger models do while frequently repeating the given executable example. In contrast, larger models like Codex and GPT-3 can generate more expressive programs with high realism, yet consequently, they often suffer from executability. We show proposed procedure can find appropriate balance and is capable of generating programs that are highly executable while maintaining reasonable expressiveness as measured by program length.

\begin{table}[H]
\centering
    \begin{tabular}{lcc}
\multicolumn{1}{c}{\bf Methods} &\multicolumn{1}{c}{\bf Executability} &\multicolumn{1}{c}{\bf Average Length} \\
\midrule
Vanilla GPT-2 1.5B & 39.40\% & 4.24 \\
Vanilla Codex 12B & 18.07\% & 7.22 \\
Vanilla GPT-3 175B & 7.79\% & 9.716 \\
\midrule
Translated Codex 12B & 78.57\% & 7.13 \\
Translated GPT-3 175B & 73.05\% & 7.36 \\
\midrule
Human & 100.00\% & 9.66
\end{tabular}
\caption{Average executability \& program length of different methods.}
\label{tab:length}
\end{table}

\section{Related Works}
Large-scale natural language modeling has witnessed rapid advances since the inception of the Transformer architecture~\citep{vaswani2017attention}. It has been shown by recent works that large language models (LLMs) pre-trained on large unstructured text corpus not only can perform strongly on various down-stream NLP tasks~\citep{devlin2018bert,radford2019language,raffel2019exploring,brown2020language} but the learned representations can also be used to model relations of entities~\citep{li2021implicit}, retrieve matching visual features~\citep{ilharco2020probing}, synthesize code from docstrings~\citep{hendrycks2021measuring,chen2021evaluating}, solve math problems~\citep{cobbe2021training,shen2021generate}, and even as valuable priors when applied to diverse tasks from different modalities~\citep{lu2021pretrained, tsimpoukelli2021multimodal}.
Notably, by pre-training on large-scale data, these models can also internalize an implicit knowledge base containing rich information about the world from which factual answers (e.g. ``Dante was born in \texttt{$\langle$PLACE$\rangle$}'') can be extracted~\citep{petroni2019language,jiang2020can,davison2019commonsense,talmor2020olmpics,roberts2020much}. Compared to prior works in \textit{single-step} knowledge extraction, we aim to extract \textit{sequential} action plans to complete open-ended human activities while satisfying various constraints of an interactive environment.

Many prior works have looked into grounding natural language in embodied environments.
A series of them parse language instructions into formal logic or rely mainly on lexical analysis to resolve various linguistic ambiguities for embodied agents~\citep{artzi2013weakly,misra2015environment,misra2016tell,tenorth2010understanding}. However, they often require many hand-designed rules or scale inadequately to more complex tasks and environments. Recently, many efforts have been put into creating more realistic environments with the goal to further advances in this area~\citep{puig2018virtualhome,shridhar2020alfred,shridhar2020alfworld,kolve2017ai2,savva2019habitat,anderson2018vision}. At the same time, by leveraging the better representation power of neural architectures, a number of works have looked into creating instruction-following agents that can perform manipulation~\citep{lynch2020grounding,lynch2021language}, navigation~\citep{fried2018speaker,wang2019reinforced,majumdar2020improving}, or both~\citep{suglia2021embodied,hill2020human,fu2019language}. Recent works also use language as hierarchical abstractions to plan actions using imitation learning~\citep{sharma2021skill} and to guide exploration in reinforcement learning~\citep{mirchandani2021ella}.

Notably, many prior works do not leverage full-blown pre-trained LLMs; most investigate smaller LMs that require considerable domain-specific data for fine-tuning to obtain reasonable performance. Perhaps more importantly, few works have evaluated LLMs in an embodiment setting that realizes the full potential of the actionable knowledge these models \textit{already contain} by pre-training on large-scale unstructured text: the tasks evaluated are often generated from a handful of templates, which do not resemble the highly diverse activities that humans perform in daily lives~\citep{harrison2016learning,jansen2020visually}.
The development of VirtualHome environment~\citep{puig2018virtualhome} enables such possibility. However, relevant works~\citep{puig2018virtualhome,liao2019synthesizing} rely on human-annotated data and perform supervised training from scratch. Due to the lack of rich world knowledge, these models can only generate action plans given detailed instructions of how to act or video demonstrations.
Concurrent work by~\citet{li2022pre} validates similar hypothesis that LMs contain rich actionable knowledge. They fine-tune GPT-2 with demonstrations to incorporate environment context and to predict actions in VirtualHome, and evaluate on tasks that are generated from pre-defined predicates. In contrast, we investigate \textit{existing} knowledge in LLMs without any additional training and evaluate on human activity tasks expressed in free-form language.

\section{Conclusion, Limitations \& Future Work}
\label{sec:conclusion}
In this work, we investigate actionable knowledge \textit{already contained} in pre-trained LLMs without any additional training. We present several techniques to extract this knowledge to perform common-sense grounding by planning actions for complex human activities.

Despite promising findings, there remain several limitations of this work which we discuss as follows:

\paragraph{Drop in Correctness}
Although our approach can significantly improve executability of the generated plans, we observe a considerable drop in correctness. In addition to the errors caused by the proposed action translation (discussed in Section~\ref{sec:translated-exec}), this is partially attributed to the limited expressivity of VirtualHome, as it may not support all necessary actions to fully complete all evaluated tasks (correctness is judged by humans). This is also reflected by that Vanilla LMs can even surpass human-written plans, which are restricted by environment expressivity.

\paragraph{Mid-Level Grounding} Instead of grounding the LLM generation to low-level actions by using downstream data from a specific environment, we focus on high-level to mid-level grounding such that we evaluate raw knowledge of LLMs as closely and broadly as possible. Hence, we only consider the most prominent challenge in mid-level grounding that the generated plans must satisfy all common-sense constraints (characterized by executability metric). As a result, we assume there is a low-level controller that can execute these mid-level actions (such as ``grab cup''), and we do not investigate the usefulness of LLMs for low-level sensorimotor behavior grounding. To perform sensorimotor grounding, such as navigation and interaction mask prediction, domain-specific data and fine-tuning are likely required.

\paragraph{Ignorant of Environment Context} We do not incorporate observation context or feedback into our models. To some extent, we approach LLMs in the same way as how VirtualHome asks human annotators to write action plans for a given human activity by \textit{imagination}, in which case humans similarly do not observe environment context. Similar to human-written plans, we assume the plans generated by LMs only refer to one instance of each object class. As a result, successful plan generation for tasks like ``stack two plates on the right side of a cup'' is not possible.

\paragraph{Evaluation Protocol} We measure quality of plans by a combination of \textit{executability} and \textit{correctness} instead of one straightforward metric. To the best of our knowledge, there isn't a known way to computationally assess the semantic correctness of the plans due to the tasks' open-ended and multi-modal nature. Prior work also adopt similar combination of metrics~\citep{puig2018virtualhome}. We report two metrics individually to shine light on the deficiencies of existing LLMs which we hope could provide insights for future works. To provide a holistic view, we report results by combining two metrics in Section~\ref{sec:grounding}.

We believe addressing each of these shortcoming will lead to exciting future directions. We also hope these findings can inspire future investigations into using pre-trained LMs for goal-driven decision-making problems and grounding the learned knowledge in embodied environments.

\section*{Acknowledgment}
We would like to thank OpenAI for providing academic access to the OpenAI API and Luke Metz for valuable feedback and discussions. This work was supported in part by Berkeley Deep Drive, NSF IIS-2024594, and GoodAI Research Award.

\newpage
\bibliography{main}
\bibliographystyle{plainnat}

\newpage
\appendix
\section{Appendix}
\subsection{Hyperparameter Search}
\label{appendix:hparam}
For each evaluated method, we perform grid search over the following hyperparameters:

\begin{table}[H]
    \centering
    \begin{tabular}{p{.2\linewidth} p{.57\linewidth} p{.23\linewidth}}
    \multicolumn{1}{c}{\bf Name} &\multicolumn{1}{c}{\bf Description} &\multicolumn{1}{c}{\bf Search Values} \\
    \midrule
    epsilon ($\epsilon$) & Out-of-distribution early termination threshold & \{0, 0.4, 0.8\} \\ \midrule
    temperature & sampling parameter adjusting relative token probabilities & \{0.1, 0.3, 0.6\} \\ \midrule
    k & number of samples generated by Planning LM & \{1, 10\} \\ \midrule
    beta ($\beta$) & weighting coefficient in action translation to trade off semantic and translation correctness & \{0.3\} \\ \midrule
    frequence\_penalty & \textbf{OpenAI API only}; penalize new tokens based on their existing frequency in the text so far & \{0.1, 0.3, 0.6, 0.9\} \\ \midrule
    presence\_penalty & \textbf{OpenAI API only}; penalize new tokens based on whether they appear in the text so far & \{0.3, 0.5, 0.8\} \\ \midrule
    repetition\_penalty & \textbf{Hugging Face Transformers only}; penalize new tokens based on whether repeating existing text & \{1.0, 1.2, 1.5, 1.8\} \\ \midrule
    \end{tabular}
\end{table}
\vspace{-.15in}

For methods that use fixed example across evaluated tasks, we search over the following three randomly chosen examples:
\begin{longtable}{p{.33\linewidth}|p{.33\linewidth}|p{.33\linewidth}}
    \bfseries Example 1 & \bfseries Example 2 & \bfseries Example 3 \\ \midrule
    Task: Use computer\newline Step 1: Walk to home office\newline Step 2: Walk to chair\newline Step 3: Find chair\newline Step 4: Sit on chair\newline Step 5: Find computer\newline Step 6: Switch on computer\newline Step 7: Turn to computer\newline Step 8: Look at computer\newline Step 9: Find keyboard\newline Step 10: Type on keyboard & Task: Relax on sofa\newline Step 1: Walk to home office\newline Step 2: Walk to couch\newline Step 3: Find couch\newline Step 4: Sit on couch\newline Step 5: Find pillow\newline Step 6: Lie on couch & Task: Read book\newline Step 1: Walk to home office\newline Step 2: Walk to novel\newline Step 3: Find novel\newline Step 4: Grab novel\newline Step 5: Find chair\newline Step 6: Sit on chair\newline Step 7: Read novel \\ \midrule
\end{longtable}

\newpage
\subsection{Details of Human Evaluations}
\label{appendix:human}
\begin{figure}[h]
\centering
\includegraphics[width=0.9\linewidth]{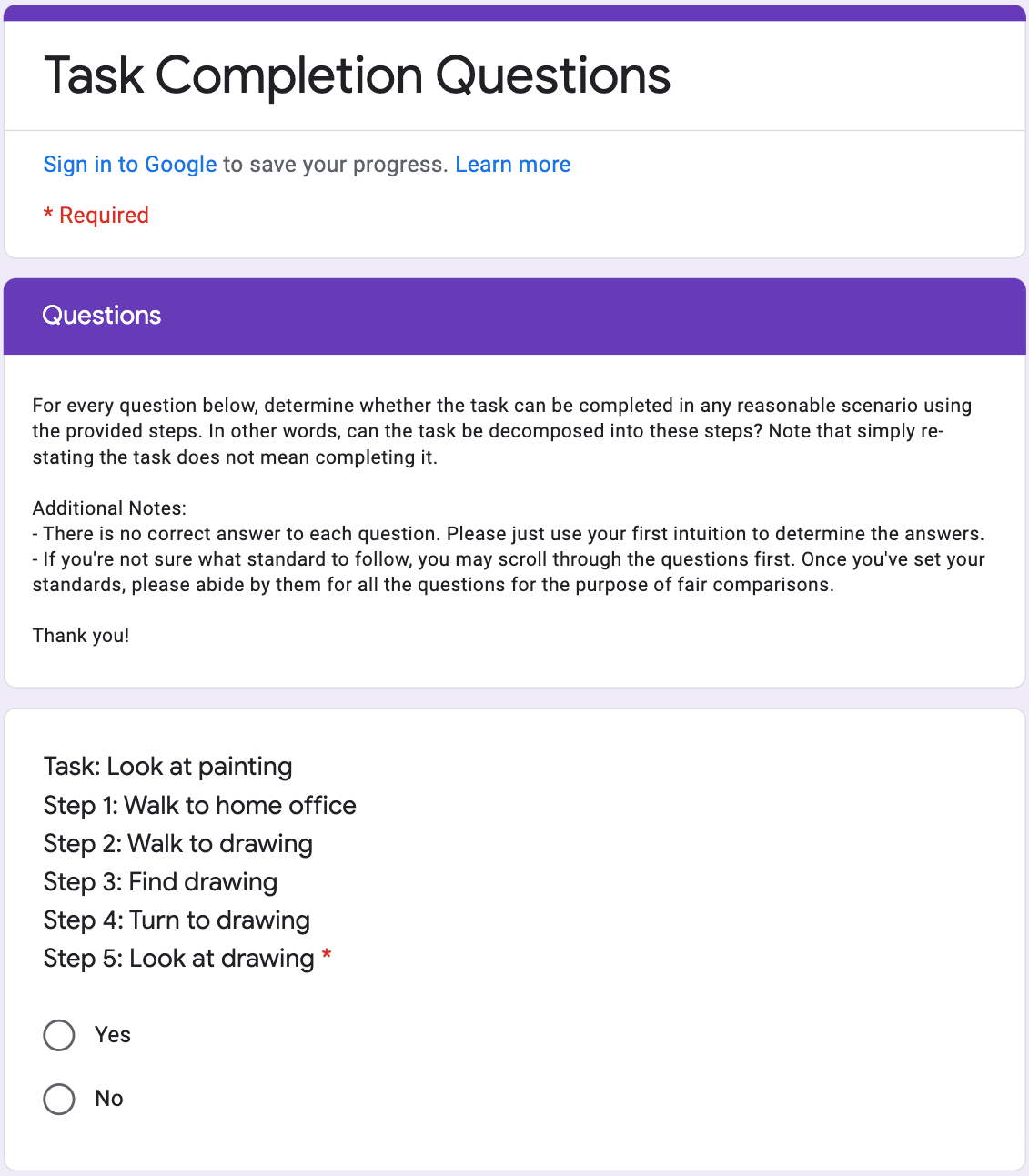}
 \caption{Screenshot of human evaluation interface, conducted as a Google Forms questionnaire.}
 \label{fig:screenshot}
\end{figure}

Human evaluations are conducted on Amazon Mechanical Turk. For each method, we generate action plans for all 88 high-level tasks. To account for the expressivity of the VirtualHome environment~\citep{puig2018virtualhome}, we include action plans written by human experts from the VirtualHome dataset as references in our human evaluations. The evaluations are conducted in the form of questionnaires containing all action plans whose order is randomly shuffled and whose corresponding methods are unknown to the annotators. Human annotators are required to answer all the questions in the questionnaire. For each question, the annotators need to answer either ``Yes'' or ``No'' indicating if they believe the action plan completes the task. For each method, we report \textit{correctness} percentage averaged across 10 participated human annotators and all 88 tasks. We further report the standard error of the mean across human annotators. Screenshot can be found in Figure~\ref{fig:screenshot}.

\newpage
\subsection{All Evaluated Tasks}

The evaluated tasks are part of the \textit{ActivityPrograms} dataset collected by~\citet{puig2018virtualhome}. Some of the task names may contain misspelling(s).

\begin{multicols}{3}
\begin{enumerate}
\item Apply lotion
\item Arrange folders
\item Breakfast
\item Browse internet
\item Brush teeth
\item Change clothes
\item Change sheets and pillow cases
\item Collect napkin rings
\item Complete surveys on amazon turk
\item Compute
\item Decorate it
\item Do homework
\item Do work
\item Draft home
\item Draw picture
\item Dry soap bottles
\item Dust
\item Eat cereal
\item Eat cheese
\item Eat snacks and drink tea
\item Empty dishwasher and fill dishwasher
\item Entertain
\item Feed me
\item Find dictionary
\item Fix snack
\item Get glass of milk
\item Give milk to cat
\item Go to sleep
\item Grab things
\item Hand washing
\item Hang keys
\item Hang pictures
\item Iron shirt
\item Keep cats inside while door is open
\item Keep cats out of room
\item Leave home
\item Listen to music
\item Look at mirror
\item Look at painting
\item Make bed
\item Make popcorn
\item Organize closet
\item Organize pantry
\item Paint ceiling
\item Pay bills
\item Pick up toys
\item Play musical chairs
\item Prepare pot of boiling water
\item Push all chairs in
\item Push in desk chair
\item Put alarm clock in bedroom
\item Put away groceries
\item Put away toys
\item Put clothes away
\item Put mail in mail organizer
\item Put on your shoes
\item Put out flowers
\item Put up decoration
\item Read
\item Read newspaper
\item Read on sofa
\item Read to child
\item Read yourself to sleep
\item Receive credit card
\item Restock
\item Scrubbing living room tile floor is once week activity for me
\item Style hair
\item Switch on lamp
\item Take jacket off
\item Take shoes off
\item Tale off shoes
\item Throw away paper
\item Try yourself off
\item Turn off TV
\item Turn on TV with remote
\item Turn on radio
\item Type up document
\item Unload various items from pockets and place them in bowl on table
\item Use laptop
\item Vacuum
\item Walk to room
\item Wash dirty dishes
\item Wash face
\item Wash monitor
\item Wash teeth
\item Watch  horror  movie
\item Wipe down sink
\item Write book
\end{enumerate}
\end{multicols}

\newpage
\subsection{Natural Language Templates for All Atomic Actions}
\label{appendix:mapping}

VirtualHome requires action steps specified in a specific format, yet language models are trained to deal with mostly natural language. We thus define a natural language template for each atomic action and only expose the converted natural language text in all operations involving language models, i.e. autoregressive generation and action translation. After we obtain an entire generated program expressed in natural language, such as those in Figure~\ref{fig:results} and Figure~\ref{fig:method}, we then convert each action step to the VirtualHome syntax. Full list of the atomic actions and their natural language templates can be found below.

\begin{table}[h]
\begin{tabular}{ll}
\multicolumn{1}{c}{\bf Atomic Action in VirtualHome Syntax} &\multicolumn{1}{c}{\bf Natural Language Template}\\
\midrule\relax
\texttt{[CLOSE] $\langle$arg1$\rangle$(1)} & close \texttt{$\langle$arg1$\rangle$}\\
\texttt{[CUT] $\langle$arg1$\rangle$(1)} & cut \texttt{$\langle$arg1$\rangle$}\\
\texttt{[DRINK] $\langle$arg1$\rangle$(1)} & drink \texttt{$\langle$arg1$\rangle$}\\
\texttt{[DROP] $\langle$arg1$\rangle$(1)} & drop \texttt{$\langle$arg1$\rangle$}\\
\texttt{[EAT] $\langle$arg1$\rangle$(1)} & eat \texttt{$\langle$arg1$\rangle$}\\
\texttt{[FIND] $\langle$arg1$\rangle$(1)} & find \texttt{$\langle$arg1$\rangle$}\\
\texttt{[GRAB] $\langle$arg1$\rangle$(1)} & grab \texttt{$\langle$arg1$\rangle$}\\
\texttt{[GREET] $\langle$arg1$\rangle$(1)} & greet \texttt{$\langle$arg1$\rangle$}\\
\texttt{[LIE] $\langle$arg1$\rangle$(1)} & lie on \texttt{$\langle$arg1$\rangle$}\\
\texttt{[LOOKAT] $\langle$arg1$\rangle$(1)} & look at \texttt{$\langle$arg1$\rangle$}\\
\texttt{[MOVE] $\langle$arg1$\rangle$(1)} & move \texttt{$\langle$arg1$\rangle$}\\
\texttt{[OPEN] $\langle$arg1$\rangle$(1)} & open \texttt{$\langle$arg1$\rangle$}\\
\texttt{[PLUGIN] $\langle$arg1$\rangle$(1)} & plug in \texttt{$\langle$arg1$\rangle$}\\
\texttt{[PLUGOUT] $\langle$arg1$\rangle$(1)} & plug out \texttt{$\langle$arg1$\rangle$}\\
\texttt{[POINTAT] $\langle$arg1$\rangle$(1)} & point at \texttt{$\langle$arg1$\rangle$}\\
\texttt{[POUR] $\langle$arg1$\rangle$(1) $\langle$arg2$\rangle$(1)} & pour \texttt{$\langle$arg1$\rangle$} into \texttt{$\langle$arg2$\rangle$}\\
\texttt{[PULL] $\langle$arg1$\rangle$(1)} & pull \texttt{$\langle$arg1$\rangle$}\\
\texttt{[PUSH] $\langle$arg1$\rangle$(1)} & push \texttt{$\langle$arg1$\rangle$}\\
\texttt{[PUTBACK] $\langle$arg1$\rangle$(1) $\langle$arg2$\rangle$(1)} & put \texttt{$\langle$arg1$\rangle$} on \texttt{$\langle$arg2$\rangle$}\\
\texttt{[PUTIN] $\langle$arg1$\rangle$(1) $\langle$arg2$\rangle$(1)} & put \texttt{$\langle$arg1$\rangle$} in \texttt{$\langle$arg2$\rangle$}\\
\texttt{[PUTOBJBACK] $\langle$arg1$\rangle$(1)} & put back \texttt{$\langle$arg1$\rangle$}\\
\texttt{[PUTOFF] $\langle$arg1$\rangle$(1)} & take off \texttt{$\langle$arg1$\rangle$}\\
\texttt{[PUTON] $\langle$arg1$\rangle$(1)} & put on \texttt{$\langle$arg1$\rangle$}\\
\texttt{[READ] $\langle$arg1$\rangle$(1)} & read \texttt{$\langle$arg1$\rangle$}\\
\texttt{[RELEASE]} & release\\
\texttt{[RINSE] $\langle$arg1$\rangle$(1)} & rinse \texttt{$\langle$arg1$\rangle$}\\
\texttt{[RUN] $\langle$arg1$\rangle$(1)} & run to \texttt{$\langle$arg1$\rangle$}\\
\texttt{[SCRUB] $\langle$arg1$\rangle$(1)} & scrub \texttt{$\langle$arg1$\rangle$}\\
\texttt{[SIT] $\langle$arg1$\rangle$(1)} & sit on \texttt{$\langle$arg1$\rangle$}\\
\texttt{[SLEEP]} & sleep\\
\texttt{[SQUEEZE] $\langle$arg1$\rangle$(1)} & squeeze \texttt{$\langle$arg1$\rangle$}\\
\texttt{[STANDUP]} & stand up\\
\texttt{[SWITCHOFF] $\langle$arg1$\rangle$(1)} & switch off \texttt{$\langle$arg1$\rangle$}\\
\texttt{[SWITCHON] $\langle$arg1$\rangle$(1)} & switch on \texttt{$\langle$arg1$\rangle$}\\
\texttt{[TOUCH] $\langle$arg1$\rangle$(1)} & touch \texttt{$\langle$arg1$\rangle$}\\
\texttt{[TURNTO] $\langle$arg1$\rangle$(1)} & turn to \texttt{$\langle$arg1$\rangle$}\\
\texttt{[TYPE] $\langle$arg1$\rangle$(1)} & type on \texttt{$\langle$arg1$\rangle$}\\
\texttt{[WAKEUP]} & wake up\\
\texttt{[WALK] $\langle$arg1$\rangle$(1)} & walk to \texttt{$\langle$arg1$\rangle$}\\
\texttt{[WASH] $\langle$arg1$\rangle$(1)} & wash \texttt{$\langle$arg1$\rangle$}\\
\texttt{[WATCH] $\langle$arg1$\rangle$(1)} & watch \texttt{$\langle$arg1$\rangle$}\\
\texttt{[WIPE] $\langle$arg1$\rangle$(1)} & wipe \texttt{$\langle$arg1$\rangle$}\\
\end{tabular}
\end{table}

\newpage
\subsection{Random Samples of Action Plans}
\label{appendix:samples}
\begin{longtable}{p{.33\linewidth}|p{.33\linewidth}|p{.33\linewidth}}
 \bfseries Human & \bfseries Vanilla GPT3-175B & \bfseries Translated GPT3-175B (Ours) \\ \midrule
Task: Go to sleep\newline Step 1: Walk to bedroom\newline Step 2: Walk to bed\newline Step 3: Find bed\newline Step 4: Turn to bed\newline Step 5: Look at bed\newline Step 6: Sit on bed\newline Step 7: Sleep & Task: Go to sleep\newline Step 1: Walk to bedroom\newline Step 2: Walk to bed\newline Step 3: Find pajamas\newline Step 4: Put on pajamas\newline Step 5: Find slippers\newline Step 6: Put on slippers\newline Step 7: Get in bed\newline Step 8: Sleep & Task: Go to sleep\newline Step 1: Walk to bedroom\newline Step 2: Walk to bed\newline Step 3: Lie on bed\newline Step 4: Sleep \\ \midrule
Task: Browse internet\newline Step 1: Walk to home office\newline Step 2: Walk to chair\newline Step 3: Find chair\newline Step 4: Sit on chair\newline Step 5: Find computer\newline Step 6: Switch on computer\newline Step 7: Turn to computer\newline Step 8: Look at computer\newline Step 9: Find keyboard\newline Step 10: Type on keyboard & Task: Browse internet\newline Step 1: Walk to home office\newline Step 2: Walk to computer\newline Step 3: Find computer\newline Step 4: Grab computer\newline Step 5: Find chair\newline Step 6: Sit on chair\newline Step 7: Browse internet & Task: Browse internet\newline Step 1: Walk to home office\newline Step 2: Walk to desk\newline Step 3: Find chair\newline Step 4: Sit on chair\newline Step 5: Find computer\newline Step 6: Switch on computer\newline Step 7: Find mouse\newline Step 8: Grab mouse\newline Step 9: Turn to computer\newline Step 10: Point at computer \\ \midrule
Task: Wash teeth\newline Step 1: Stand up\newline Step 2: Walk to bathroom\newline Step 3: Walk to sink\newline Step 4: Find sink\newline Step 5: Turn to sink\newline Step 6: Look at sink\newline Step 7: Find teeth\newline Step 8: Wash teeth & Task: Wash teeth\newline Step 1: Walk to bathroom\newline Step 2: Walk to sink\newline Step 3: Turn on faucet\newline Step 4: Grab toothbrush\newline Step 5: Put toothpaste on toothbrush\newline Step 6: Put toothbrush in mouth\newline Step 7: Brush teeth\newline Step 8: Rinse mouth out with water\newline Step 9: Turn off faucet\newline Step 10: Walk back to bedroom & Task: Wash teeth\newline Step 1: Walk to bathroom\newline Step 2: Walk to sink\newline Step 3: Find faucet\newline Step 4: Switch on faucet\newline Step 5: Walk to toothbrush\newline Step 6: Grab toothbrush \\ \midrule
Task: Brush teeth\newline Step 1: Walk to bathroom\newline Step 2: Walk to tooth paste\newline Step 3: Find tooth paste\newline Step 4: Grab tooth paste\newline Step 5: Find toothbrush\newline Step 6: Grab toothbrush & Task: Brush teeth\newline Step 1: Walk to bathroom\newline Step 2: Walk to sink\newline Step 3: Turn on water\newline Step 4: Grab toothbrush\newline Step 5: Put toothpaste on toothbrush\newline Step 6: Put toothbrush in mouth\newline Step 7: Brush teeth & Task: Brush teeth\newline Step 1: Find toothbrush\newline Step 2: Turn to toothbrush\newline Step 3: Point at toothbrush\newline Step 4: Wash toothbrush \\ \midrule
Task: Vacuum\newline Step 1: Walk to home office\newline Step 2: Walk to computer\newline Step 3: Find vacuum cleaner\newline Step 4: Switch on vacuum cleaner\newline Step 5: Switch off vacuum cleaner & Task: Vacuum\newline Step 1: Walk to home office\newline Step 2: Walk to vacuum\newline Step 3: Find vacuum\newline Step 4: Grab vacuum\newline Step 5: Find chair\newline Step 6: Sit on chair\newline Step 7: Vacuum & Task: Vacuum\newline Step 1: Walk to home office\newline Step 2: Walk to dresser\newline Step 3: Open dresser\newline Step 4: Find vacuum cleaner\newline Step 5: Grab vacuum cleaner\newline Step 6: Pull vacuum cleaner\newline Step 7: Plug in vacuum cleaner\newline Step 8: Switch on vacuum cleaner \\ \midrule
Task: Change sheets and pillow cases\newline Step 1: Find pillow\newline Step 2: Turn to pillow\newline Step 3: Look at pillow\newline Step 4: Grab pillow & Task: Change sheets and pillow cases\newline Step 1: Walk to bedroom\newline Step 2: Walk to bed\newline Step 3: Change sheets and pillow cases\newline Step 4: Walk to kitchen\newline Step 5: Walk to sink\newline Step 6: Wash sheets and pillow cases\newline Step 7: Walk to bedroom\newline Step 8: Put sheets and pillow cases on bed\newline Step 9: Walk to living room\newline Step 10: Sit on couch & Task: Change sheets and pillow cases\newline Step 1: Walk to bedroom\newline Step 2: Walk to bed\newline Step 3: Find sheets\newline Step 4: Grab sheets\newline Step 5: Find pillow\newline Step 6: Grab pillow \\ \midrule
Task: Wash dirty dishes\newline Step 1: Walk to dining room\newline Step 2: Walk to sink\newline Step 3: Find sponge\newline Step 4: Grab sponge\newline Step 5: Find faucet\newline Step 6: Switch on faucet\newline Step 7: Put sponge on sink\newline Step 8: Grab sponge\newline Step 9: Find dish soap\newline Step 10: Grab dish soap\newline Step 11: Pour dish soap into sponge\newline Step 12: Put back dish soap\newline Step 13: Find plate\newline Step 14: Grab plate\newline Step 15: Scrub plate\newline Step 16: Rinse plate\newline Step 17: Find dishrack\newline Step 18: Put plate on dishrack\newline Step 19: Find plate\newline Step 20: Grab plate\newline Step 21: Scrub plate\newline Step 22: Rinse plate\newline Step 23: Put plate on dishrack\newline Step 24: Rinse sponge\newline Step 25: Put back sponge\newline Step 26: Switch off faucet & Task: Wash dirty dishes\newline Step 1: Walk to kitchen\newline Step 2: Walk to sink\newline Step 3: Wash dishes\newline Step 4: Dry dishes\newline Step 5: Put dishes away & Task: Wash dirty dishes\newline Step 1: Walk to dining room\newline Step 2: Walk to sink\newline Step 3: Find faucet\newline Step 4: Switch on faucet\newline Step 5: Walk to plate\newline Step 6: Grab plate \\ \midrule
Task: Feed me\newline Step 1: Walk to dining room\newline Step 2: Walk to plate\newline Step 3: Find plate\newline Step 4: Grab plate\newline Step 5: Find food\newline Step 6: Grab food\newline Step 7: Put food on plate\newline Step 8: Find food\newline Step 9: Grab food\newline Step 10: Put food on plate\newline Step 11: Find fork\newline Step 12: Grab fork\newline Step 13: Eat plate\newline Step 14: Eat plate & Task: Feed me\newline Step 1: Walk to kitchen\newline Step 2: Walk to refrigerator\newline Step 3: Open refrigerator\newline Step 4: Grab food\newline Step 5: Find plate\newline Step 6: Put food on plate\newline Step 7: Find fork\newline Step 8: Put fork on plate\newline Step 9: Find spoon\newline Step 10: Put spoon on plate\newline Step 11: Find cup\newline Step 12: Pour water in cup\newline Step 13: Walk to table\newline Step 14: Sit at table\newline Step 15: Eat food & Task: Feed me\newline Step 1: Walk to dining room\newline Step 2: Walk to table\newline Step 3: Find table\newline Step 4: Turn to table\newline Step 5: Find chair\newline Step 6: Sit on chair\newline Step 7: Find food\newline Step 8: Grab food\newline Step 9: Find plate\newline Step 10: Put food on plate \\ \midrule
Task: Breakfast\newline Step 1: Walk to dining room\newline Step 2: Walk to food\newline Step 3: Grab food\newline Step 4: Walk to bedroom\newline Step 5: Walk to television\newline Step 6: Switch on television\newline Step 7: Walk to bed\newline Step 8: Lie on bed\newline Step 9: Turn to television\newline Step 10: Watch television\newline Step 11: Eat food & Task: Breakfast\newline Step 1: Walk to kitchen\newline Step 2: Find cereal\newline Step 3: Pour cereal into bowl\newline Step 4: Pour milk into bowl\newline Step 5: Walk to living room\newline Step 6: Sit on couch\newline Step 7: Eat cereal & Task: Breakfast\newline Step 1: Walk to dining room\newline Step 2: Walk to freezer\newline Step 3: Find freezer\newline Step 4: Open freezer\newline Step 5: Find bread\newline Step 6: Grab bread\newline Step 7: Close freezer\newline Step 8: Find toaster\newline Step 9: Plug in toaster\newline Step 10: Put bread on plate\newline Step 11: Switch on toaster \\ \midrule
Task: Read\newline Step 1: Find novel\newline Step 2: Grab novel\newline Step 3: Walk to bedroom\newline Step 4: Walk to chair\newline Step 5: Sit on chair\newline Step 6: Read novel & Task: Read\newline Step 1: Walk to home office\newline Step 2: Walk to novel\newline Step 3: Find novel\newline Step 4: Grab novel\newline Step 5: Find chair\newline Step 6: Sit on chair\newline Step 7: Read novel & Task: Read\newline Step 1: Walk to home office\newline Step 2: Walk to light\newline Step 3: Find light\newline Step 4: Switch on light\newline Step 5: Find novel\newline Step 6: Grab novel\newline Step 7: Find chair\newline Step 8: Sit on chair\newline Step 9: Read novel \\ \midrule
Task: Style hair\newline Step 1: Find towel\newline Step 2: Turn to towel\newline Step 3: Point at towel\newline Step 4: Wash towel\newline Step 5: Turn to towel\newline Step 6: Look at towel\newline Step 7: Squeeze towel\newline Step 8: Walk to bathroom & Task: Style hair\newline Step 1: Walk to bathroom\newline Step 2: Walk to mirror\newline Step 3: Find hairbrush\newline Step 4: Grab hairbrush\newline Step 5: Find chair\newline Step 6: Sit on chair\newline Step 7: Style hair & Task: Style hair\newline Step 1: Walk to bathroom\newline Step 2: Walk to mirror\newline Step 3: Find hair\newline Step 4: Grab hair\newline Step 5: Find comb\newline Step 6: Grab comb\newline Step 7: Find brush\newline Step 8: Grab brush \\ \midrule
Task: Pick up toys\newline Step 1: Walk to bedroom\newline Step 2: Walk to toy\newline Step 3: Find toy\newline Step 4: Grab toy\newline Step 5: Walk to filing cabinet\newline Step 6: Open filing cabinet\newline Step 7: Put toy on filing cabinet\newline Step 8: Find toy\newline Step 9: Grab toy\newline Step 10: Walk to filing cabinet\newline Step 11: Put toy on filing cabinet\newline Step 12: Find toy\newline Step 13: Grab toy\newline Step 14: Walk to filing cabinet\newline Step 15: Put toy on filing cabinet\newline Step 16: Close filing cabinet & Task: Pick up toys\newline Step 1: Walk to living room\newline Step 2: Walk to toy box\newline Step 3: Find toy box\newline Step 4: Grab toy box\newline Step 5: Find chair\newline Step 6: Sit on chair\newline Step 7: Pick up toys & Task: Pick up toys\newline Step 1: Find toy\newline Step 2: Turn to toy\newline Step 3: Point at toy\newline Step 4: Wash toy \\ \midrule
Task: Decorate it\newline Step 1: Walk to home office\newline Step 2: Walk to centerpiece\newline Step 3: Find centerpiece\newline Step 4: Grab centerpiece\newline Step 5: Walk to table\newline Step 6: Put centerpiece on table\newline Step 7: Find drawing\newline Step 8: Grab drawing\newline Step 9: Find wall\newline Step 10: Turn to wall\newline Step 11: Put drawing on wall\newline Step 12: Find drawing\newline Step 13: Grab drawing\newline Step 14: Turn to wall\newline Step 15: Put drawing on wall & Task: Decorate it\newline Step 1: Walk to home office\newline Step 2: Walk to novel\newline Step 3: Find novel\newline Step 4: Grab novel\newline Step 5: Find chair\newline Step 6: Sit on chair\newline Step 7: Read novel\newline Step 8: Decorate it & Task: Decorate it\newline Step 1: Walk to table\newline Step 2: Walk to centerpiece\newline Step 3: Look at centerpiece\newline Step 4: Grab centerpiece\newline Step 5: Point at centerpiece \\ \midrule
Task: Put away groceries\newline Step 1: Walk to dining room\newline Step 2: Walk to freezer\newline Step 3: Find freezer\newline Step 4: Open freezer\newline Step 5: Find food\newline Step 6: Grab food\newline Step 7: Put food in freezer\newline Step 8: Close freezer & Task: Put away groceries\newline Step 1: Walk to kitchen\newline Step 2: Walk to refrigerator\newline Step 3: Open refrigerator\newline Step 4: Grab milk and eggs\newline Step 5: Close refrigerator\newline Step 6: Walk to counter\newline Step 7: Put milk and eggs on counter\newline Step 8: Walk to cabinet\newline Step 9: Open cabinet\newline Step 10: Put away milk and eggs\newline Step 11: Close cabinet\newline Step 12: Walk to refrigerator\newline Step 13: Grab orange juice and bread\newline Step 14: Close refrigerator & Task: Put away groceries\newline Step 1: Walk to kitchen\newline Step 2: Walk to fridge\newline Step 3: Open fridge\newline Step 4: Put food in fridge\newline Step 5: Close fridge\newline Step 6: Walk to dining room\newline Step 7: Walk to livingroom\newline Step 8: Walk to bedroom\newline Step 9: Walk to bathroom \\ \midrule
Task: Draw picture\newline Step 1: Walk to home office\newline Step 2: Walk to chair\newline Step 3: Find chair\newline Step 4: Sit on chair\newline Step 5: Find desk\newline Step 6: Open desk\newline Step 7: Find check\newline Step 8: Grab check\newline Step 9: Put check on desk\newline Step 10: Find pencil\newline Step 11: Grab pencil\newline Step 12: Close desk & Task: Draw picture\newline Step 1: Walk to home office\newline Step 2: Walk to pencils\newline Step 3: Find pencils\newline Step 4: Grab pencils\newline Step 5: Find paper\newline Step 6: Sit on chair\newline Step 7: Draw picture & Task: Draw picture\newline Step 1: Walk to home office\newline Step 2: Walk to wall\newline Step 3: Walk to drawing\newline Step 4: Find drawing\newline Step 5: Turn to drawing\newline Step 6: Look at drawing\newline Step 7: Touch drawing\newline Step 8: Pull drawing\newline Step 9: Push drawing \\ \midrule
Task: Entertain\newline Step 1: Walk to home office\newline Step 2: Walk to cheese\newline Step 3: Find cheese\newline Step 4: Find food\newline Step 5: Find couch\newline Step 6: Sit on couch\newline Step 7: Find wine glass\newline Step 8: Grab wine glass\newline Step 9: Drink wine glass & Task: Entertain\newline Step 1: Walk to living room\newline Step 2: Walk to entertainment center\newline Step 3: Find remote\newline Step 4: Grab remote\newline Step 5: Find chair\newline Step 6: Sit on chair\newline Step 7: Entertain self & Task: Entertain\newline Step 1: Walk to livingroom\newline Step 2: Walk to couch\newline Step 3: Find couch\newline Step 4: Sit on couch \\ \midrule
Task: Type up document\newline Step 1: Walk to home office\newline Step 2: Walk to chair\newline Step 3: Find chair\newline Step 4: Sit on chair\newline Step 5: Find computer\newline Step 6: Switch on computer\newline Step 7: Turn to computer\newline Step 8: Look at computer\newline Step 9: Find check\newline Step 10: Grab check\newline Step 11: Read check\newline Step 12: Find coffee cup\newline Step 13: Grab coffee cup\newline Step 14: Drink coffee cup\newline Step 15: Put back coffee cup\newline Step 16: Find keyboard\newline Step 17: Type on keyboard & Task: Type up document\newline Step 1: Walk to home office\newline Step 2: Walk to keyboard\newline Step 3: Find keyboard\newline Step 4: Grab keyboard\newline Step 5: Find chair\newline Step 6: Sit on chair\newline Step 7: Type document & Task: Type up document\newline Step 1: Walk to home office\newline Step 2: Walk to computer\newline Step 3: Find computer\newline Step 4: Switch on computer\newline Step 5: Find chair\newline Step 6: Sit on chair\newline Step 7: Turn to computer\newline Step 8: Look at computer\newline Step 9: Find keyboard\newline Step 10: Grab keyboard\newline Step 11: Type on keyboard\newline Step 12: Stand up\newline Step 13: Find printer\newline Step 14: Switch on printer\newline Step 15: Find printing paper \\ \midrule
Task: Put on your shoes\newline Step 1: Walk to home office\newline Step 2: Walk to shoes\newline Step 3: Find shoes\newline Step 4: Grab shoes\newline Step 5: Find chair\newline Step 6: Sit on chair\newline Step 7: Find feet both\newline Step 8: Put shoes on feet both & Task: Put on your shoes\newline Step 1: Walk to home office\newline Step 2: Walk to shoes\newline Step 3: Find shoes\newline Step 4: Grab shoes\newline Step 5: Find chair\newline Step 6: Sit on chair\newline Step 7: Put on shoes & Task: Put on your shoes\newline Step 1: Find shoes\newline Step 2: Turn to shoes\newline Step 3: Point at shoes\newline Step 4: Wash shoes \\ \midrule
Task: Watch  horror  movie\newline Step 1: Walk to home office\newline Step 2: Walk to remote control\newline Step 3: Find remote control\newline Step 4: Grab remote control\newline Step 5: Find couch\newline Step 6: Sit on couch\newline Step 7: Find television\newline Step 8: Switch on television\newline Step 9: Turn to television\newline Step 10: Watch television & Task: Watch  horror  movie\newline Step 1: Walk to home office\newline Step 2: Walk to horror movie\newline Step 3: Find horror movie\newline Step 4: Grab horror movie\newline Step 5: Find chair\newline Step 6: Sit on chair\newline Step 7: Watch horror movie & Task: Watch  horror  movie\newline Step 1: Find television\newline Step 2: Switch on television\newline Step 3: Find cd player\newline Step 4: Switch on cd player\newline Step 5: Switch off cd player\newline Step 6: Find remote control\newline Step 7: Switch on remote control\newline Step 8: Switch off remote control \\ \midrule
Task: Turn on radio\newline Step 1: Walk to home office\newline Step 2: Walk to stereo\newline Step 3: Find stereo\newline Step 4: Find electrical outlet\newline Step 5: Turn to electrical outlet\newline Step 6: Plug in stereo\newline Step 7: Switch on electrical outlet\newline Step 8: Switch on stereo & Task: Turn on radio\newline Step 1: Walk to home office\newline Step 2: Walk to radio\newline Step 3: Find radio\newline Step 4: Grab radio\newline Step 5: Find chair\newline Step 6: Sit on chair\newline Step 7: Turn on radio & Task: Turn on radio\newline Step 1: Walk to home office\newline Step 2: Walk to chair\newline Step 3: Find chair\newline Step 4: Sit on chair\newline Step 5: Find remote control\newline Step 6: Grab remote control\newline Step 7: Find radio\newline Step 8: Point at radio\newline Step 9: Switch on radio\newline Step 10: Put back remote control \\ \midrule
Task: Look at mirror\newline Step 1: Walk to bathroom\newline Step 2: Walk to mirror\newline Step 3: Find mirror\newline Step 4: Turn to mirror\newline Step 5: Look at mirror & Task: Look at mirror\newline Step 1: Walk to bathroom\newline Step 2: Walk to mirror\newline Step 3: Look in mirror & Task: Look at mirror\newline Step 1: Walk to bedroom\newline Step 2: Walk to mirror\newline Step 3: Find mirror\newline Step 4: Turn to mirror\newline Step 5: Look at mirror\newline Step 6: Find face\newline Step 7: Turn to face\newline Step 8: Look at face\newline Step 9: Find hair\newline Step 10: Turn to hair\newline Step 11: Look at hair\newline Step 12: Find dress\newline Step 13: Turn to dress\newline Step 14: Look at dress \\ \midrule
Task: Leave home\newline Step 1: Walk to home office\newline Step 2: Walk to shoes\newline Step 3: Find shoes\newline Step 4: Grab shoes\newline Step 5: Put on shoes & Task: Leave home\newline Step 1: Walk to front door\newline Step 2: Open front door\newline Step 3: Walk to car\newline Step 4: Open car door\newline Step 5: Get in car\newline Step 6: Close car door\newline Step 7: Drive to work & Task: Leave home\newline Step 1: Find shoes\newline Step 2: Grab shoes\newline Step 3: Put on shoes \\ \midrule
Task: Walk to room\newline Step 1: Walk to home office\newline Step 2: Walk to home office & Task: Walk to room\newline Step 1: Walk to home office\newline Step 2: Walk to novel\newline Step 3: Find novel\newline Step 4: Grab novel\newline Step 5: Find chair\newline Step 6: Sit on chair\newline Step 7: Read novel & Task: Walk to room\newline Step 1: Walk to dining room\newline Step 2: Walk to livingroom\newline Step 3: Walk to bedroom \\ \midrule
Task: Hang pictures\newline Step 1: Walk to bedroom\newline Step 2: Walk to drawing\newline Step 3: Find drawing\newline Step 4: Grab drawing\newline Step 5: Find wall\newline Step 6: Put drawing on wall\newline Step 7: Turn to drawing\newline Step 8: Look at drawing\newline Step 9: Find drawing\newline Step 10: Grab drawing\newline Step 11: Put drawing on wall\newline Step 12: Turn to drawing\newline Step 13: Look at drawing & Task: Hang pictures\newline Step 1: Walk to home office\newline Step 2: Walk to box of pictures\newline Step 3: Find box of pictures\newline Step 4: Grab box of pictures\newline Step 5: Walk to wall\newline Step 6: Hang pictures on wall & Task: Hang pictures\newline Step 1: Walk to home office\newline Step 2: Walk to table\newline Step 3: Find keys\newline Step 4: Grab keys\newline Step 5: Walk to wall\newline Step 6: Find hanger\newline Step 7: Put keys on hanger \\ \midrule
Task: Wipe down sink\newline Step 1: Walk to dining room\newline Step 2: Walk to rag\newline Step 3: Find rag\newline Step 4: Grab rag\newline Step 5: Walk to sink\newline Step 6: Wipe sink\newline Step 7: Put back rag & Task: Wipe down sink\newline Step 1: Walk to kitchen\newline Step 2: Walk to sink\newline Step 3: Grab sponge\newline Step 4: Wipe down sink\newline Step 5: Rinse sponge\newline Step 6: Put sponge away\newline Step 7: Wash hands & Task: Wipe down sink\newline Step 1: Walk to bathroom\newline Step 2: Walk to cleaning solution\newline Step 3: Find cleaning solution\newline Step 4: Grab cleaning solution\newline Step 5: Find sink\newline Step 6: Pour cleaning solution into sink\newline Step 7: Find faucet\newline Step 8: Switch on faucet\newline Step 9: Scrub sink\newline Step 10: Wash sink \\ \midrule
Task: Take shoes off\newline Step 1: Walk to home office\newline Step 2: Walk to shoes\newline Step 3: Find shoes\newline Step 4: Find shoes\newline Step 5: Take off shoes\newline Step 6: Find dresser\newline Step 7: Open dresser\newline Step 8: Grab shoes\newline Step 9: Put shoes in dresser\newline Step 10: Close dresser & Task: Take shoes off\newline Step 1: Walk to home office\newline Step 2: Walk to shoes\newline Step 3: Find shoes\newline Step 4: Grab shoes\newline Step 5: Find chair\newline Step 6: Sit on chair\newline Step 7: Take shoes off & Task: Take shoes off\newline Step 1: Walk to home office\newline Step 2: Walk to shoes\newline Step 3: Find shoes \\ \midrule
Task: Give milk to cat\newline Step 1: Walk to home office\newline Step 2: Walk to cat\newline Step 3: Find cat\newline Step 4: Turn to cat\newline Step 5: Point at cat\newline Step 6: Pull cat\newline Step 7: Find milk\newline Step 8: Grab milk\newline Step 9: Find plate\newline Step 10: Pour milk into plate\newline Step 11: Touch cat\newline Step 12: Scrub cat & Task: Give milk to cat\newline Step 1: Walk to kitchen\newline Step 2: Walk to fridge\newline Step 3: Open fridge\newline Step 4: Grab milk\newline Step 5: Walk to cat\newline Step 6: Pour milk into bowl\newline Step 7: Put bowl on floor\newline Step 8: Walk away from cat & Task: Give milk to cat\newline Step 1: Find cat\newline Step 2: Touch cat\newline Step 3: Open fridge\newline Step 4: Touch milk\newline Step 5: Pour milk into bowl \\ \midrule
Task: Grab things\newline Step 1: Walk to novel\newline Step 2: Turn to novel\newline Step 3: Grab novel & Task: Grab things\newline Step 1: Walk to home office\newline Step 2: Walk to novel\newline Step 3: Find novel\newline Step 4: Grab novel\newline Step 5: Find chair\newline Step 6: Sit on chair\newline Step 7: Read novel\newline Step 8: Grab things & Task: Grab things\newline Step 1: Find toy\newline Step 2: Turn to toy\newline Step 3: Point at toy\newline Step 4: Wash toy \\ \midrule
Task: Use laptop\newline Step 1: Walk to home office\newline Step 2: Walk to couch\newline Step 3: Find couch\newline Step 4: Sit on couch\newline Step 5: Find laptop\newline Step 6: Switch on laptop\newline Step 7: Turn to laptop\newline Step 8: Look at laptop\newline Step 9: Find keyboard\newline Step 10: Type on keyboard\newline Step 11: Turn to keyboard\newline Step 12: Look at keyboard\newline Step 13: Type on keyboard\newline Step 14: Turn to laptop\newline Step 15: Look at laptop & Task: Use laptop\newline Step 1: Walk to home office\newline Step 2: Walk to laptop\newline Step 3: Find laptop\newline Step 4: Grab laptop\newline Step 5: Find chair\newline Step 6: Sit on chair\newline Step 7: Use laptop & Task: Use laptop\newline Step 1: Walk to home office\newline Step 2: Walk to chair\newline Step 3: Find chair\newline Step 4: Sit on chair\newline Step 5: Find laptop\newline Step 6: Switch on laptop\newline Step 7: Find mouse\newline Step 8: Grab mouse\newline Step 9: Find mousepad\newline Step 10: Put mouse on mousepad\newline Step 11: Turn to laptop\newline Step 12: Point at laptop \\ \midrule
Task: Organize pantry\newline Step 1: Walk to dining room\newline Step 2: Walk to pantry\newline Step 3: Find food\newline Step 4: Grab food\newline Step 5: Turn to food\newline Step 6: Look at food\newline Step 7: Find garbage can\newline Step 8: Put food on garbage can\newline Step 9: Find vegetable\newline Step 10: Grab vegetable\newline Step 11: Turn to vegetable\newline Step 12: Look at vegetable\newline Step 13: Put vegetable on garbage can\newline Step 14: Find dry pasta\newline Step 15: Grab dry pasta\newline Step 16: Turn to dry pasta\newline Step 17: Look at dry pasta\newline Step 18: Put back dry pasta\newline Step 19: Find food\newline Step 20: Turn to food\newline Step 21: Look at food\newline Step 22: Push food\newline Step 23: Find noodles\newline Step 24: Grab noodles\newline Step 25: Turn to noodles\newline Step 26: Look at noodles\newline Step 27: Put back noodles & Task: Organize pantry\newline Step 1: Walk to kitchen\newline Step 2: Walk to pantry\newline Step 3: Find pantry\newline Step 4: Grab pantry\newline Step 5: Find chair\newline Step 6: Sit on chair\newline Step 7: Organize pantry & Task: Organize pantry\newline Step 1: Walk to pantry\newline Step 2: Find pantry \\ \midrule
Task: Change clothes\newline Step 1: Walk to bedroom\newline Step 2: Walk to dresser\newline Step 3: Open dresser\newline Step 4: Find dress\newline Step 5: Grab dress\newline Step 6: Put on dress\newline Step 7: Close dresser & Task: Change clothes\newline Step 1: Walk to bedroom\newline Step 2: Walk to closet\newline Step 3: Find clothes\newline Step 4: Grab clothes\newline Step 5: Find chair\newline Step 6: Sit on chair\newline Step 7: Change clothes & Task: Change clothes\newline Step 1: Walk to bedroom\newline Step 2: Walk to dresser\newline Step 3: Find dresser\newline Step 4: Open dresser\newline Step 5: Find pants\newline Step 6: Grab pants\newline Step 7: Put on pants\newline Step 8: Find shirt\newline Step 9: Grab shirt\newline Step 10: Put on shirt\newline Step 11: Close dresser \\ \midrule
Task: Put up decoration\newline Step 1: Walk to home office\newline Step 2: Walk to drawing\newline Step 3: Find drawing\newline Step 4: Find centerpiece\newline Step 5: Grab centerpiece\newline Step 6: Walk to wall\newline Step 7: Grab drawing\newline Step 8: Put drawing on wall\newline Step 9: Find centerpiece\newline Step 10: Grab centerpiece\newline Step 11: Walk to table\newline Step 12: Put centerpiece on table & Task: Put up decoration\newline Step 1: Walk to home office\newline Step 2: Walk to decoration\newline Step 3: Find decoration\newline Step 4: Grab decoration\newline Step 5: Find chair\newline Step 6: Sit on chair\newline Step 7: Put up decoration & Task: Put up decoration\newline Step 1: Walk to livingroom\newline Step 2: Walk to table \\ \midrule
Task: Make bed\newline Step 1: Walk to bedroom\newline Step 2: Walk to bed\newline Step 3: Find sheets\newline Step 4: Grab sheets\newline Step 5: Pull sheets & Task: Make bed\newline Step 1: Walk to bedroom\newline Step 2: Walk to bed\newline Step 3: Find sheets\newline Step 4: Find blanket\newline Step 5: Find pillow\newline Step 6: Put sheets on bed\newline Step 7: Put blanket on bed\newline Step 8: Put pillow on bed & Task: Make bed\newline Step 1: Stand up\newline Step 2: Find sheets\newline Step 3: Find blanket\newline Step 4: Find pillow\newline Step 5: Find comforter\newline Step 6: Find bed \\ \midrule
Task: Scrubbing living room tile floor is once week activity for me\newline Step 1: Find mop\newline Step 2: Turn to mop\newline Step 3: Point at mop\newline Step 4: Wash mop\newline Step 5: Turn to mop\newline Step 6: Look at mop & Task: Scrubbing living room tile floor is once week activity for me\newline Step 1: Walk to living room\newline Step 2: Walk to tile floor\newline Step 3: Find scrub brush\newline Step 4: Grab scrub brush\newline Step 5: Scrub tile floor\newline Step 6: Put down scrub brush\newline Step 7: Walk to kitchen & Task: Scrubbing living room tile floor is once week activity for me\newline Step 1: Walk to livingroom\newline Step 2: Walk to mop\newline Step 3: Find mop\newline Step 4: Grab mop\newline Step 5: Walk to bathroom\newline Step 6: Find bucket\newline Step 7: Switch on faucet \\ \midrule
Task: Read yourself to sleep\newline Step 1: Walk to bedroom\newline Step 2: Walk to novel\newline Step 3: Find novel\newline Step 4: Grab novel\newline Step 5: Find bed\newline Step 6: Sit on bed\newline Step 7: Find bookmark\newline Step 8: Grab bookmark\newline Step 9: Read novel\newline Step 10: Put bookmark on novel\newline Step 11: Sleep & Task: Read yourself to sleep\newline Step 1: Walk to bedroom\newline Step 2: Walk to bed\newline Step 3: Find book\newline Step 4: Grab book\newline Step 5: Find chair\newline Step 6: Sit on chair\newline Step 7: Read book\newline Step 8: Sleep & Task: Read yourself to sleep\newline Step 1: Walk to bedroom\newline Step 2: Walk to bed\newline Step 3: Lie on bed\newline Step 4: Sleep\newline Step 5: Read book \\ \midrule
Task: Hand washing\newline Step 1: Walk to bathroom\newline Step 2: Walk to sink\newline Step 3: Find faucet\newline Step 4: Switch on faucet\newline Step 5: Find soap\newline Step 6: Grab soap\newline Step 7: Find hands both\newline Step 8: Scrub hands both\newline Step 9: Put back soap\newline Step 10: Wash hands both\newline Step 11: Switch off faucet\newline Step 12: Find towel\newline Step 13: Grab towel\newline Step 14: Wipe hands both & Task: Hand washing\newline Step 1: Walk to bathroom\newline Step 2: Walk to sink\newline Step 3: Turn on faucet\newline Step 4: Grab soap\newline Step 5: Wash hands\newline Step 6: Rinse hands\newline Step 7: Dry hands & Task: Hand washing\newline Step 1: Walk to bathroom\newline Step 2: Walk to sink\newline Step 3: Find faucet\newline Step 4: Turn to faucet\newline Step 5: Find hands both\newline Step 6: Wash hands both \\ \midrule
Task: Throw away paper\newline Step 1: Walk to home office\newline Step 2: Walk to desk\newline Step 3: Find desk\newline Step 4: Turn to desk\newline Step 5: Find chair\newline Step 6: Sit on chair\newline Step 7: Find check\newline Step 8: Grab check\newline Step 9: Squeeze check\newline Step 10: Stand up\newline Step 11: Walk to trashcan\newline Step 12: Put check on trashcan & Task: Throw away paper\newline Step 1: Walk to home office\newline Step 2: Walk to wastebasket\newline Step 3: Find wastebasket\newline Step 4: Grab wastebasket\newline Step 5: Walk to desk\newline Step 6: Drop paper in wastebasket & Task: Throw away paper\newline Step 1: Walk to home office\newline Step 2: Walk to table\newline Step 3: Find table\newline Step 4: Turn to table\newline Step 5: Find paper\newline Step 6: Grab paper\newline Step 7: Walk to trashcan\newline Step 8: Open trashcan\newline Step 9: Put paper on trashcan\newline Step 10: Close trashcan \\ \midrule
Task: Tale off shoes\newline Step 1: Walk to home office\newline Step 2: Walk to shoes\newline Step 3: Find shoes\newline Step 4: Find shoes\newline Step 5: Take off shoes\newline Step 6: Grab shoes\newline Step 7: Find mat\newline Step 8: Put shoes on mat & Task: Tale off shoes\newline Step 1: Walk to home office\newline Step 2: Walk to novel\newline Step 3: Find novel\newline Step 4: Grab novel\newline Step 5: Find chair\newline Step 6: Sit on chair\newline Step 7: Read novel & Task: Tale off shoes\newline Step 1: Walk to home office\newline Step 2: Walk to shoes\newline Step 3: Find shoes \\ \midrule
Task: Hang keys\newline Step 1: Walk to bedroom\newline Step 2: Walk to dresser\newline Step 3: Find purse\newline Step 4: Grab purse\newline Step 5: Open purse\newline Step 6: Find keys\newline Step 7: Grab keys\newline Step 8: Find bowl\newline Step 9: Put keys on bowl\newline Step 10: Close purse\newline Step 11: Put back purse & Task: Hang keys\newline Step 1: Walk to home office\newline Step 2: Walk to key hook\newline Step 3: Find key hook\newline Step 4: Grab key hook\newline Step 5: Find keys\newline Step 6: Hang keys on key hook\newline Step 7: Walk to front door\newline Step 8: Leave house & Task: Hang keys\newline Step 1: Walk to home office\newline Step 2: Walk to table\newline Step 3: Find keys\newline Step 4: Grab keys\newline Step 5: Walk to wall\newline Step 6: Find hanger\newline Step 7: Put keys on hanger \\ \midrule
Task: Make popcorn\newline Step 1: Walk to dining room\newline Step 2: Walk to food\newline Step 3: Find food\newline Step 4: Grab food\newline Step 5: Walk to microwave\newline Step 6: Turn to microwave\newline Step 7: Open microwave\newline Step 8: Put food in microwave\newline Step 9: Close microwave\newline Step 10: Switch on microwave & Task: Make popcorn\newline Step 1: Walk to kitchen\newline Step 2: Walk to microwave\newline Step 3: Open microwave\newline Step 4: Put bag of popcorn in microwave\newline Step 5: Close microwave\newline Step 6: Push buttons on microwave\newline Step 7: Wait for popcorn to pop\newline Step 8: Get bowl out of cupboard\newline Step 9: Get butter out of fridge\newline Step 10: Get salt out of cupboard\newline Step 11: Pour popcorn into bowl\newline Step 12: Add butter and salt to bowl\newline Step 13: & Task: Make popcorn\newline Step 1: Walk to kitchen\newline Step 2: Walk to microwave\newline Step 3: Find microwave\newline Step 4: Open microwave \\ \midrule
Task: Receive credit card\newline Step 1: Walk to home office\newline Step 2: Walk to envelope\newline Step 3: Find envelope\newline Step 4: Grab envelope\newline Step 5: Walk to home office\newline Step 6: Walk to scissors\newline Step 7: Find scissors\newline Step 8: Open envelope\newline Step 9: Find creditcard\newline Step 10: Grab creditcard & Task: Receive credit card\newline Step 1: Walk to mailbox\newline Step 2: Open mailbox\newline Step 3: Grab credit card\newline Step 4: Walk to home office\newline Step 5: Walk to desk\newline Step 6: Put credit card on desk\newline Step 7: Sit at desk\newline Step 8: Look at credit card & Task: Receive credit card\newline Step 1: Walk to home office\newline Step 2: Walk to cabinet\newline Step 3: Find cabinet\newline Step 4: Open cabinet \\ \midrule
Task: Read to child\newline Step 1: Walk to bedroom\newline Step 2: Walk to child\newline Step 3: Find child\newline Step 4: Greet child\newline Step 5: Walk to filing cabinet\newline Step 6: Open filing cabinet\newline Step 7: Find novel\newline Step 8: Grab novel\newline Step 9: Walk to chair\newline Step 10: Pull chair\newline Step 11: Sit on chair\newline Step 12: Turn to novel\newline Step 13: Look at novel\newline Step 14: Turn to novel\newline Step 15: Point at novel\newline Step 16: Turn to novel\newline Step 17: Point at novel\newline Step 18: Read novel\newline Step 19: Read novel\newline Step 20: Read novel & Task: Read to child\newline Step 1: Walk to home office\newline Step 2: Walk to novel\newline Step 3: Find novel\newline Step 4: Grab novel\newline Step 5: Find chair\newline Step 6: Sit on chair\newline Step 7: Read novel & Task: Read to child\newline Step 1: Walk to child\newline Step 2: Read book \\ \midrule
Task: Eat cheese\newline Step 1: Walk to dining room\newline Step 2: Walk to freezer\newline Step 3: Open freezer\newline Step 4: Find cheese\newline Step 5: Grab cheese\newline Step 6: Eat cheese & Task: Eat cheese\newline Step 1: Walk to kitchen\newline Step 2: Walk to refrigerator\newline Step 3: Open refrigerator\newline Step 4: Grab cheese\newline Step 5: Eat cheese & Task: Eat cheese\newline Step 1: Walk to dining room\newline Step 2: Walk to table\newline Step 3: Find table\newline Step 4: Turn to table\newline Step 5: Find chair\newline Step 6: Sit on chair\newline Step 7: Find food\newline Step 8: Grab food\newline Step 9: Find plate\newline Step 10: Put food on plate \\ \midrule
Task: Unload various items from pockets and place them in bowl on table\newline Step 1: Walk to home office\newline Step 2: Walk to table\newline Step 3: Find bowl\newline Step 4: Grab bowl\newline Step 5: Put bowl on table\newline Step 6: Find pants\newline Step 7: Find bills\newline Step 8: Grab bills\newline Step 9: Find phone\newline Step 10: Grab phone & Task: Unload various items from pockets and place them in bowl on table\newline Step 1: Walk to home office\newline Step 2: Walk to bowl on table\newline Step 3: Find bowl on table\newline Step 4: Grab bowl on table\newline Step 5: Unload various items from pockets and place them in bowl on table & Task: Unload various items from pockets and place them in bowl on table\newline Step 1: Find bowl\newline Step 2: Turn to bowl\newline Step 3: Point at bowl\newline Step 4: Wash bowl \\ \midrule
Task: Switch on lamp\newline Step 1: Walk to bedroom\newline Step 2: Walk to floor lamp\newline Step 3: Find floor lamp\newline Step 4: Switch on floor lamp & Task: Switch on lamp\newline Step 1: Walk to lamp\newline Step 2: Switch on lamp\newline Step 3: Walk to chair\newline Step 4: Sit on chair\newline Step 5: Read book & Task: Switch on lamp\newline Step 1: Walk to bedroom\newline Step 2: Walk to floor lamp\newline Step 3: Find floor lamp\newline Step 4: Turn to floor lamp\newline Step 5: Switch on floor lamp\newline Step 6: Find bed\newline Step 7: Lie on bed \\ \midrule
Task: Take jacket off\newline Step 1: Walk to bedroom\newline Step 2: Walk to jacket\newline Step 3: Find jacket\newline Step 4: Find jacket\newline Step 5: Take off jacket\newline Step 6: Find dresser\newline Step 7: Open dresser\newline Step 8: Close dresser & Task: Take jacket off\newline Step 1: Walk to home office\newline Step 2: Walk to jacket\newline Step 3: Find jacket\newline Step 4: Grab jacket\newline Step 5: Find chair\newline Step 6: Sit on chair\newline Step 7: Take jacket off & Task: Take jacket off\newline Step 1: Walk to home office\newline Step 2: Walk to jacket\newline Step 3: Find jacket
\end{longtable}

\end{document}